\documentclass[final,12pt,authoryear]{elsarticle}

\usepackage{amssymb}
\usepackage{amsmath}
\usepackage{kotex}
\usepackage{url}
\usepackage{subfig}
\usepackage{color, colortbl}
\usepackage{booktabs}
\usepackage{microtype}
\definecolor{lightblue}{rgb}{0.886, 0.929, 0.996}
\usepackage{lineno}
\usepackage[colorlinks=true,linkcolor=blue,citecolor=blue,urlcolor=blue]{hyperref}

\usepackage{titlesec}

\titleformat{\subsection}
  {\normalfont\bfseries\normalsize}  
  {\thesubsection}{1em}{}       
  
\titleformat{\paragraph}
  {\normalfont\bfseries\normalsize}
  {\theparagraph}{1em}{}

\journal{arXiv}

\begin{document}

\begin{frontmatter}

\title{Enhancing Zero-shot Commonsense Reasoning by Integrating Visual Knowledge via Machine Imagination} %% Article title

\author[label1]{Hyuntae Park}
\ead{pht0639@korea.ac.kr}

\author[label1]{Yeachan Kim}
\ead{yeachan@korea.ac.kr}

\author[label1,label2]{SangKeun Lee}
\ead{yalphy@korea.ac.kr}

\affiliation[label1]{organization={Department of Artificial Intelligence, Korea University},
            addressline={145 Anam-ro, Seungbuk-gu},
            city={Seoul},
            postcode={02841}, 
            % state={},
            country={Republic of Korea}}

\affiliation[label2]{organization={Department of Computer Science and Engineering, Korea University},
            addressline={145 Anam-ro, Seungbuk-gu},
            city={Seoul},
            postcode={02841}, 
            % state={},
            country={Republic of Korea}}

%% Abstract
\begin{abstract}
Recent advancements in zero-shot commonsense reasoning have empowered Pre-trained Language Models (PLMs) to acquire extensive commonsense knowledge without requiring task-specific fine-tuning. Despite this progress, these models frequently suffer from limitations caused by human reporting biases inherent in textual knowledge, leading to understanding discrepancies between machines and humans. To bridge this gap, we introduce an additional modality to enrich the reasoning capabilities of PLMs.
We propose \textsc{Imagine} (Machine \textbf{Imagin}ation-based R\textbf{e}asoning), a novel zero-shot commonsense reasoning framework that supplements textual inputs with visual signals from machine-generated images. Specifically, we enhance PLMs with the ability to \textbf{imagine} by embedding an image generator directly into the reasoning pipeline. To facilitate effective utilization of this imagined visual context, we construct synthetic datasets designed to emulate visual question-answering scenarios.
Through comprehensive evaluations on multiple commonsense reasoning benchmarks, we demonstrate that \textsc{Imagine} substantially outperforms existing zero-shot approaches and even surpasses advanced large language models. These results underscore the capability of machine imagination to mitigate reporting bias and significantly enhance the generalization ability of commonsense reasoning models. Our code and data are available at \url{https://github.com/Park-ing-lot/Imagine}.
\end{abstract}

%%Graphical abstract
% \begin{graphicalabstract}
%\includegraphics{grabs}
% \end{graphicalabstract}

%%Research highlights
% \begin{highlights}
% \item Introduce a new framework for zero-shot commonsense reasoning.
% \item Leverage machine-generated images to enhance language understanding.
% \item Exceed existing state-of-the-art methods in zero-shot reasoning tasks.
% \item Reduce reporting bias with machine-generated visual knowledge.
% \end{highlights}

%% Keywords
\begin{keyword}
Natural Language Processing \sep Commonsense Reasoning \sep Multi-modal \sep Zero-shot Inference
%% keywords here, in the form: keyword \sep keyword

%% PACS codes here, in the form: \PACS code \sep code

%% MSC codes here, in the form: \MSC code \sep code
%% or \MSC[2008] code \sep code (2000 is the default)

\end{keyword}

\end{frontmatter}
%% Add \usepackage{lineno} before \begin{document} and uncomment 
%% following line to enable line numbers
%% \linenumbers

%% main text
% \linenumbers
\section{Introduction}
Commonsense reasoning is regarded as a key milestone in advancing artificial general intelligence \citep{gunning2018machine}. Although Pre-trained Language Models (PLMs; \citet{DBLP:conf/nips/VaswaniSPUJGKP17, reimers-2019-sentence-bert, DBLP:conf/nips/BrownMRSKDNSSAA20-gpt}) demonstrate nearly human-level reasoning abilities after fine-tuning on specific commonsense datasets, they still struggle in zero-shot scenarios where examples deviate significantly from their training data distribution \citep{DBLP:journals/corr/abs-1909-08855-mitra, DBLP:conf/naacl/KimKKAHY22, YANG2025106816}. Addressing this limitation is essential to achieving more comprehensive natural language understanding at a human-like level.

A promising approach to overcoming this challenge involves augmenting PLMs with commonsense knowledge drawn from external Knowledge Bases (KBs; \citet{DBLP:conf/aaai/SapBABLRRSC19-atomic, DBLP:conf/aaai/HwangBBDSBC21-atomic2020, DBLP:journals/corr/abs-2206-01532-abstractatomic}). Specifically, this method converts knowledge entities into a question-answering (QA) format, creating a synthetic QA dataset. This dataset is then used to train PLMs in a manner similar to their initial pre-training phase. As these knowledge bases encompass a wide range of commonsense information, this approach improves reasoning abilities across various situations without needing to specialize in specific knowledge areas \citep{DBLP:conf/emnlp/WangF0XLSB23-car, DBLP:journals/corr/abs-2401-07286-candle}.

\begin{figure}[t!]
\centering
  \includegraphics[width=\columnwidth]{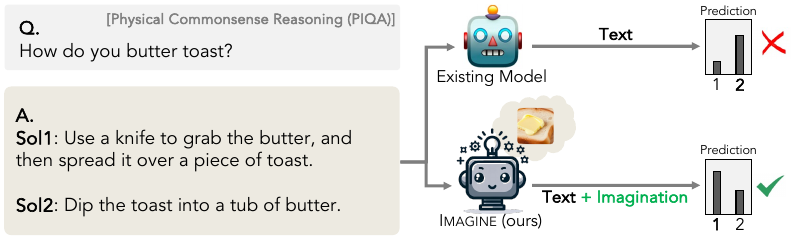}
    \caption{An example from the PIQA dataset \citep{DBLP:conf/aaai/BiskZLGC20-piqa} with model predictions. \textsc{Imagine} performs reasoning by leveraging machine-generated images to enhance understanding of the question.}
    \label{fig:fig1}
\end{figure}

However, PLMs often face limitations from human reporting bias \citep{DBLP:conf/cikm/GordonD13-reporting}, as commonsense knowledge in text form typically emphasizes the most common scenarios, overlooking less frequent but equally critical knowledge needed for holistic reasoning. Figure \ref{fig:fig1} highlights a situation where a recent model \citep{DBLP:conf/emnlp/WangF0XLSB23-car} struggles to accurately respond to the question "\textit{How do you butter toast?}". Since these models rely solely on textual inputs, they often miss contextual nuances, like the fact that butter is generally too solid to dip. In contrast, humans can easily address such questions by visualizing the texture, solidity, and interactions of butter with other objects. This limitation motivates the exploration of additional modalities to enrich textual commonsense knowledge.

To address this, we introduce \textsc{Imagine} (Machine \textbf{Imagin}ation-based R\textbf{e}asoning), a novel zero-shot commonsense reasoning framework designed to mitigate reporting biases inherent in textual information.
Drawing inspiration from cognitive studies highlighting the benefits of visual imagery in language understanding \citep{Gambrell1986MentalIA, Dessalegn2013InteractionBL}, \textsc{Imagine} enhances PLMs with additional visual signals generated from the input question to complement textual input.
To equip PLMs with this machine imagination capability, we incorporate either conditional image generation models \citep{betker2023improving-dalle3, YANG2025125583} or vision-language models capable of retrieving semantically relevant images \citep{DBLP:conf/icml/RadfordKHRGASAM21-clip}.
Additionally, we construct Synthetic VQA and Synthetic VQA$\boldsymbol{+}$, datasets consisting of commonsense knowledge represented in a question–answer format accompanied by visual content.
By training on these datasets, \textsc{Imagine} learns to jointly interpret language and imagery, thereby expanding its accessible commonsense knowledge and reducing its reliance on human-reported textual biases.

To assess the effectiveness of \textsc{Imagine}, we conduct extensive experiments spanning various reasoning benchmarks, architectures, and model scales. Results from these experiments clearly show that \textsc{Imagine} outperforms existing models, including large language models, in reasoning capabilities. Additionally, our detailed analysis reveals that \textsc{Imagine} enables PLMs to dynamically leverage machine imagination in beneficial ways.

This study serves as an extension of our preliminary work \citep{park2024zero}, with three key advancements.
First, we enhance our core Synthetic VQA dataset by introducing Synthetic VQA$\boldsymbol{+}$, which incorporates a broader spectrum of visual commonsense knowledge to cover more diverse scenarios. To further improve knowledge quality, we filter out implausible elements that do not meaningfully support visual imagination-based reasoning, thereby maximizing the model’s ability to leverage visual signals.

Second, to mitigate the inefficiency of generating new images at inference time—a limitation of conventional PLMs enhanced with visual imagination—we propose an alternative approach that utilizes retrieved images. We also discuss the practical benefits this method brings in terms of efficiency and performance.
Lastly, we conduct comprehensive experiments incorporating all extended components, and our results demonstrate consistent improvements over the initial framework. This culminates in new state-of-the-art performance on zero-shot commonsense reasoning tasks and validates the effectiveness of our proposed extensions.

The primary contributions of this study are as follows:
\begin{itemize}
    % \item We present \textsc{Imagine}, a novel framework for zero-shot commonsense reasoning designed to reduce reporting bias and improve the generalization abilities of PLMs.
    
    % \item We develop the Synthetic VQA$\boldsymbol{+}$ dataset, enabling PLMs to integrate both textual and visual inputs effectively for enhanced commonsense reasoning.
    
    % \item We show that \textsc{Imagine} outperforms state-of-the-art zero-shot reasoning models across diverse tasks, underscoring the importance of machine imagination.
    \item We present \textsc{Imagine}, a novel framework for zero-shot commonsense reasoning that addresses reporting bias and improves the generalization abilities of PLMs. This work demonstrates that incorporating visual information into text-based commonsense reasoning can meaningfully enhance reasoning performance.

    \item We introduce \textit{Synthetic VQA} and \textit{Synthetic VQA$\boldsymbol{+}$}, new multimodal datasets that provide visual–textual commonsense knowledge. These resources are designed to enhance PLMs’ imagination capabilities for improved text-based commonsense reasoning.
    
    \item \textsc{Imagine} achieves state-of-the-art performance on a range of zero-shot commonsense reasoning benchmarks, outperforming even much larger models such as GPT-4, despite being built on language models with fewer than 1B parameters.
\end{itemize}

\section{Related Work}
In this section, we review prior work on zero-shot commonsense reasoning and the use of visual signals in natural language understanding. We also discuss recent advances in multimodal large language models and multimodal commonsense knowledge bases, and clarify how our approach differs from these lines of research.

\subsection{Zero-shot Commonsense Reasoning}
Zero-shot commonsense reasoning is generally approached in two main ways. The first approach leverages the inherent knowledge embedded within off-the-shelf PLMs without any parameter updates. For instance, \citet{DBLP:journals/corr/abs-1806-02847} measured language model perplexity as a proxy for commonsense reasoning, while \citet{DBLP:conf/emnlp/LiKHdBN22} used specially crafted prompts to enhance PLM's performance. \citet{DBLP:conf/emnlp/ShwartzWBBC20-selftalk} introduced an iterative self-talk process to elicit commonsense knowledge from the models. Similarly, \citet{DBLP:conf/aaai/DouP22} applied cloze-style translation to extract additional knowledge for reasoning. These methods rely on querying or prompting to draw out latent commonsense knowledge from the models without the need for fine-tuning.

The second approach involves leveraging external commonsense knowledge bases (e.g., ATOMIC \citep{DBLP:conf/aaai/SapBABLRRSC19-atomic}, ConceptNet \citep{DBLP:conf/aaai/SpeerCH17-conceptnet}) to provide language models with additional knowledge. Specifically, recent studies have transformed the knowledge entities (e.g., triplets of (head, relation, tail)) into synthetic QA pairs and trained the models with them \citep{DBLP:conf/emnlp/BanerjeeB20-smlm,DBLP:conf/aaai/MaIFBNO21-ma}. Recently, \citet{DBLP:conf/emnlp/WangF0XLSB23-car} introduced a conceptualization method \citep{song2011short} that generalizes commonsense triplets to a range of higher-level instances. Subsequently, \citet{DBLP:journals/corr/abs-2401-07286-candle} incorporated an instantiation phase in the synthetic data generation process, leveraging the generative abilities of large language models (LLMs).
These knowledge-enhanced approaches broaden the commonsense scope accessible to language models.

While these works have advanced zero-shot performance on commonsense benchmarks, they overlook reporting bias \citep{DBLP:conf/cikm/GordonD13-reporting}, potentially impeding an accurate understanding of commonsense knowledge. In contrast, our method helps pre-trained language models sidestep inaccurate reasoning by augmenting their capabilities with visual imagination. By infusing visual commonsense knowledge, our approach addresses limitations left by text-centric methods, allowing the model to tackle zero-shot commonsense reasoning tasks with a multimodal perspective.

\subsection{Incorporating Visual Signals for Natural Language Understanding}

Some prior research has explored the use of machine imagination for tackling Natural Language Understanding (NLU) tasks. For example, \citet{tan2020vokenization} proposed VOKEN, which introduces visual supervision into language model pre-training by incorporating external knowledge from images retrieved for the tokens. Instead of retrieving visual information, \citet{lu2022imagination} proposed generating synthetic images (i.e., imagination) based on a generative model to tackle downstream NLU tasks. In the context of commonsense reasoning, \citet{liu2022things} utilized visual information to comprehend spatial commonsense knowledge (e.g., \textit{how big is a lion?}). Similar to the proposed method, \citet{DBLP:conf/emnlp/YangYZWYC22-zlavi} introduced Z-LaVI, which integrated visual information with PLMs through both retrieval and synthesis to achieve zero-shot reasoning abilities. 
Unlike these approaches that apply visual signals directly, our method introduces a distinct pre-training phase, enabling the model to develop a baseline of visual imagination that can be generalized across zero-shot reasoning tasks.

\subsection{Vision-Language Models and Commonsense Knowledge Bases}
Recent advances in multimodal models \citep{DBLP:conf/nips/Dai0LTZW0FH23-instructblip, DBLP:conf/nips/LiuLWL23a, DBLP:conf/iclr/Zhu0SLE24-minigpt4, DBLP:journals/corr/abs-2408-08872-blip3} have attempted to transfer the success of language models to vision–language tasks by grounding language in perceptual inputs. These models typically focus on reasoning about physical entities or contextual cues within real images, emphasizing grounded understanding in visually observable environments \citep{DBLP:conf/nips/Huang0WHSML0MPL23-kosmos-grounding, WU2025107593}. Moreover, most of these systems rely heavily on static, high-quality images and are designed to solve tasks that are inherently visual in nature.

In contrast, our approach adopts a language-centered perspective: we aim to enhance text-based reasoning by incorporating machine imagination—often incomplete or abstract visual signals generated or retrieved by the model itself. Rather than grounding language in vision, we reinterpret vision as a tool to amplify the model's linguistic capacity. Our goal is to solve language tasks by transforming them into vision–language reasoning tasks, where visual inputs are integrated in a way that supports rather than dominates the underlying language understanding.

Similarly, prior visual commonsense knowledge bases \citep{DBLP:conf/cvpr/ZellersBFC19-vcr, DBLP:conf/eccv/ParkBMFC20, DBLP:conf/eccv/HesselHPZBRSC22} have been constructed with strong reliance on visual context, making the associated tasks unsolvable without images. In our framework, however, tasks are primarily language-based and solvable through textual reasoning. Visual signals are selectively integrated to supplement reasoning in cases where textual inputs suffer from reporting bias or underspecification.

\section{Machine Imagination-based Reasoning}

\begin{figure*}[t]
\centering
\subfloat[Construction procedures of Synthetic VQA$\boldsymbol{+}$ dataset]{%
  \includegraphics[width=\linewidth]{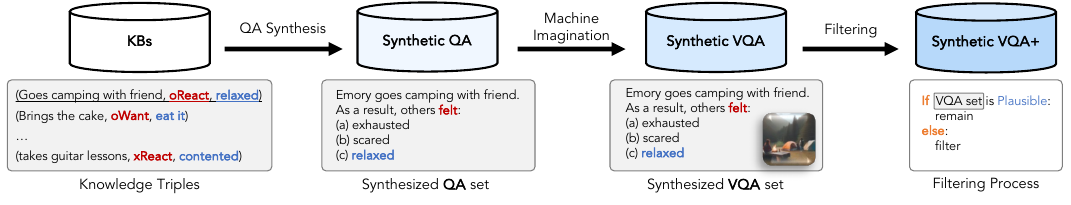}%
  % \label{exp:mot:acc}
}

\subfloat[Inference and optimization procedures of \textsc{Imagine} (ours)]{%
  \includegraphics[width=\linewidth]{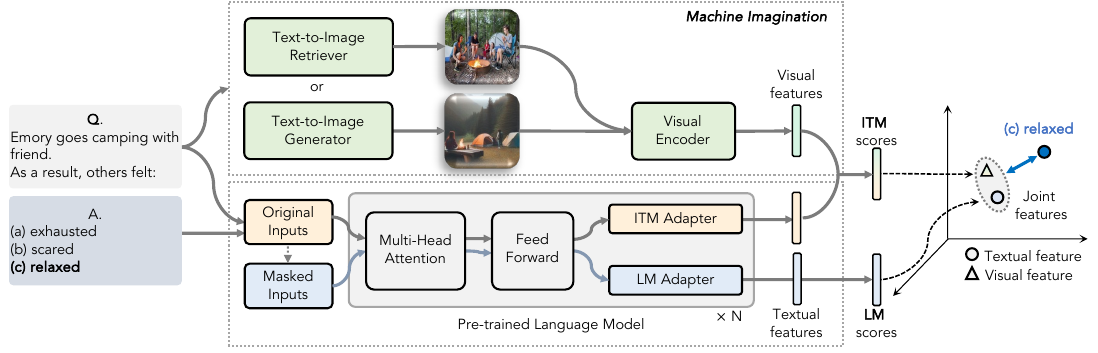}%
  % \label{exp:mot:acc}
}% \hfill
\caption{Overall procedures for (a) constructing a Synthetic VQA dataset and (b) the inference/optimization phase of \textsc{Imagine} (ours) using the given QA pair. The process starts with the textual pair consisting of a question and its answers, followed by the generation of visual signals (i.e., imagination) conditioned on the question. The two distinct features from visual and textual models are then utilized to derive a comprehensive prediction.}
\label{fig:model}
\end{figure*}

\begin{figure*}[t!]
\centering
  \includegraphics[width=\columnwidth]{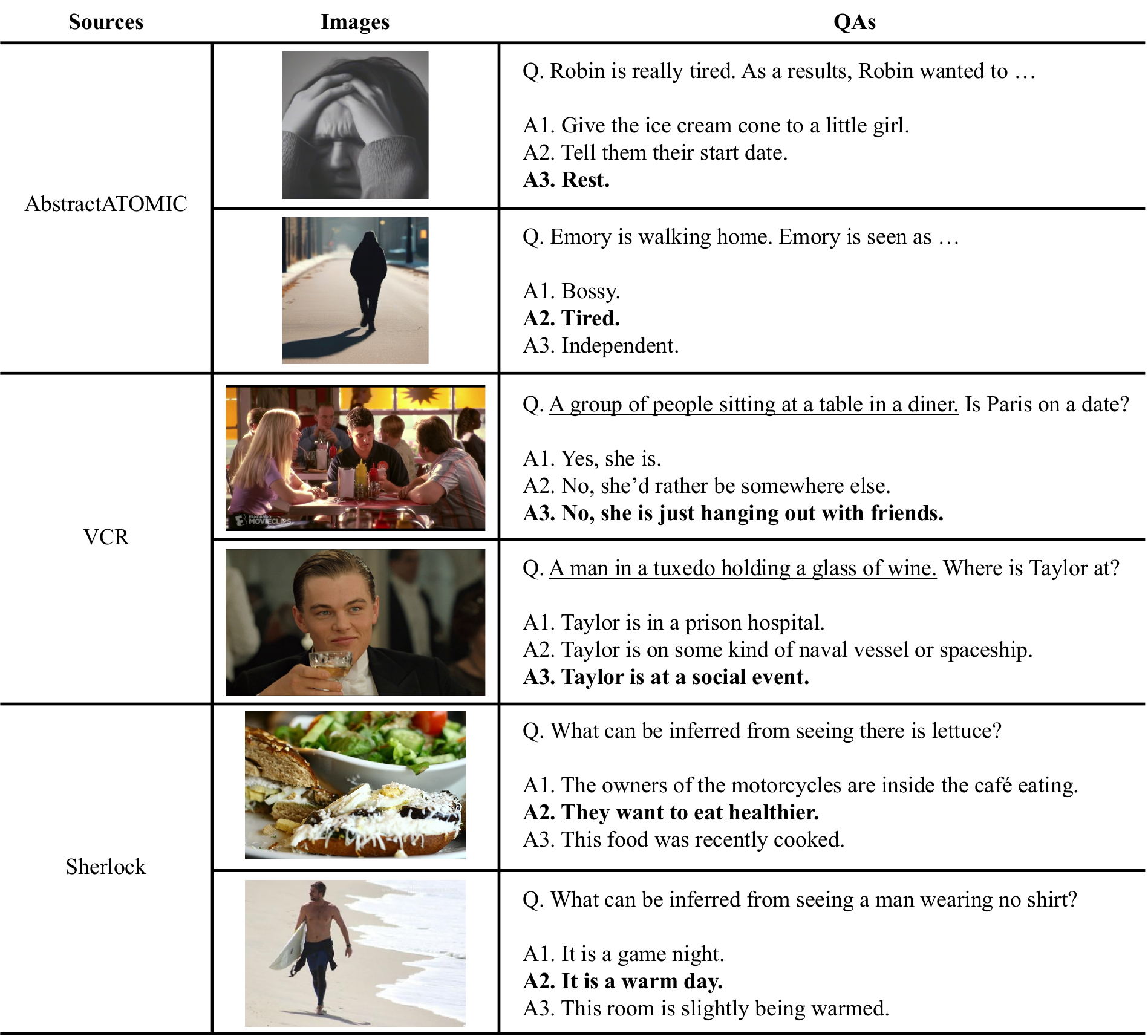}
    \caption{Examples of the Synthetic VQA$\boldsymbol{+}$ dataset. Our dataset is sourced from AbstractATOMIC \citep{DBLP:conf/emnlp/WangF0XLSB23-car}, VCR \citep{DBLP:conf/cvpr/ZellersBFC19-vcr}, and Sherlock \citep{DBLP:conf/eccv/HesselHPZBRSC22}. \textbf{Bold} indicates the correct answer, and \underline{Underline} denotes the generated image caption.}
    \label{fig:vqa_examples}
\end{figure*}

In this section, we elaborate on the proposed method, namely \textsc{Imagine} (Machine \textbf{Imagin}ation-based R\textbf{e}asoning), for zero-shot commonsense reasoning.  The core strategy is to complement textual commonsense knowledge with visual signals derived from machine-generated or retirved images. To achieve this, we first couple the PLMs with a text-to-image generator (\S\ref{subsec:machine_imagination}), enabling machine imagination in text-based PLMs. We then construct a large-scale Synthetic VQA series to learn the joint use of textual and visual signals in the reasoning process (\S\ref{subsec:vqa}). By optimizing the model with additional signals that encapsulate commonsense knowledge, \textsc{Imagine} can effectively perform commonsense reasoning while avoiding human reporting bias inherent in textual inputs (\S\ref{subsec:injection}, \S\ref{subsec:inference}). The overall procedure is depicted in Figure \ref{fig:model}.

\subsection{Machine Imagination in PLMs}\label{subsec:machine_imagination}
We begin by integrating machine imagination into text-based PLMs. Let $M_{T}$ denote a Pre-trained Language Model, which forms the foundation for zero-shot commonsense reasoning. To enable machine imagination, we enhance this model with two additional visual modules: (i) a text-to-image generator $M_{T2I}$, responsible for generating relevant images conditioned on textual inputs, and (ii) a visual encoder $M_{I}$, which extracts informative visual features from the generated images.

The overall workflow of machine imagination proceeds as follows. Given a textual input, the text-to-image model $M_{T2I}$ first synthesizes an image that visually represents the textual description. Subsequently, the textual PLM $M_{T}$ and visual encoder $M_{I}$ collaboratively encode both the textual input and the generated visual information. The integrated representation obtained from these models is then leveraged to produce enhanced predictions for commonsense reasoning tasks.

\subsection{Synthetic VQA \& Synthetic VQA$\boldsymbol{+}$ Construction}\label{subsec:vqa}
Following the previous works \citep{DBLP:conf/aaai/MaIFBNO21-ma,DBLP:conf/emnlp/WangF0XLSB23-car}, we achieve zero-shot commonsense reasoning ability by constructing the synthetic QA dataset from the knowledge base. On top of this dataset, we build a synthetic visual question-answering (Synthetic VQA) dataset with the help of machine imagination. Additionally, we incorporate a visual commonsense dataset that contains real images \citep{DBLP:conf/cvpr/ZellersBFC19-vcr, DBLP:conf/eccv/HesselHPZBRSC22}. The dataset is designed to: (i) instill commonsense reasoning abilities in PLMs and (ii) teach them to harmoniously utilize both textual and visual inputs. Examples of our dataset can be found in Figure \ref{fig:vqa_examples}.

The objective of this process is to construct VQA pairs $(Q, A, I)$, where each pair includes a natural language question $Q$, a set of $n$ answer choices $A={A_1, A_2, ..., A_n}$, including one ground-truth answer and $n-1$ distractors, along with an image $I$ that corresponds to the question.

\paragraph{Synthetic VQA} 
We first construct textual QA pairs from the KBs by following the recent work \citep{DBLP:conf/emnlp/WangF0XLSB23-car}. Specifically, we transform the knowledge entities into the QA pairs through the conceptualized augmentation of the entities \citep{DBLP:conf/emnlp/WangF0XLSB23-car} with the pre-defined natural language templates (e.g., the relation of \textit{xWant} is transformed to \textit{As a result, PersonX wanted to}). This process results in textual synthetic QA pairs $(Q, A)$.
On the textual synthetic QA pairs, we input the textual question $Q$ to the text-to-image model $M_{T2I}$ to generate the visual counterpart $I$ that depicts the scenarios described in each question. In this process, we standardize all person names in the questions to “Person” and remove duplicate questions. These generated images provide an additional layer of information, offering a visual context that enhances the reasoning ability based not only on textual descriptions but also on visual evidence. This augmentation leverages the strengths of visual imagery on language understanding \citep{Gambrell1986MentalIA,Dessalegn2013InteractionBL}, potentially improving the robustness and accuracy of the model predictions.

However, relying solely on the synthetic relationships between QA pairs and generated images can introduce challenges in aligning visual content accurately since machines often fail to generate well-aligned images with textual inputs \citep{DBLP:conf/iclr/FengHFJANBWW23-t2i_alignment}.
% \footnote{Additional analysis related to this alignment is provided in the experiments section.}. 
Therefore, we supplement the Synthetic VQA pairs with the widely used Visual Commonsense Reasoning (VCR) dataset \citep{DBLP:conf/cvpr/ZellersBFC19-vcr}. Each pair in this dataset consists of $(Q, A, R, I)$, where $R$ is a rationale for the correct answer. We omit $R$ since our focus is on the QA pairs associated with relevant images. Additionally, to enrich the input and enhance visual comprehension for PLMs, we generate textual context for each image using an image captioning model, such as InstructBLIP \citep{DBLP:conf/nips/Dai0LTZW0FH23-instructblip}, and prepend this context as a prefix to each $Q$. The statistics for Synthetic VQA are provided in Table \ref{tab:statistics}.

\begin{table}[t!]
\centering
\scriptsize
\begin{tabular}{@{}l|cc|c@{}}
\toprule
 & Train & Dev & Total \\ \midrule
\# Images generated from AbsAT & 18,838 & 1,695 & 20,533 \\
\# QA pairs from AbsAT & 486,778 & 46,238 & 533,016 \\ \midrule
\# Images from VCR & 80,418 & 9,929 & 90,347 \\
\# QA pairs from VCR & 212,923 & 26,534 & 239,457 \\ \midrule
\# Total Images & 99,256 & 11,624 & 110,880 \\
\# Total QA pairs & 699,701 & 72,772 & 772,473 \\ \bottomrule
\end{tabular}%
\caption{Statistic of Synthetic VQA dataset.}
\label{tab:statistics}
\end{table}

\paragraph{Synthetic VQA$\boldsymbol{+}$} 
In this extended study, we aim to enhance Synthetic VQA by improving its quality to examine how diverse and plausible visual commonsense knowledge impacts visual imagination. To achieve this, we introduce two key enhancements: augmenting the dataset with richer visual commonsense information and filtering out implausible samples for higher data reliability.

To enrich Synthetic VQA, we incorporate data from Sherlock \citep{DBLP:conf/eccv/HesselHPZBRSC22}, a visual commonsense reasoning dataset designed to capture various everyday scenarios. Sherlock provides annotations that pair depicted details (called clues) with inferred commonsense conclusions (called inferences), thus offering a diverse range of visual commonsense knowledge. Specifically, we transform $(\text{clue}, \text{inference}, I)$ triples into $(Q, A, I)$ format to align with the Synthetic VQA structure. To further enhance the challenge for models, we identify distractors by calculating the cosine similarity between inferences using SentenceBERT embeddings \citep{reimers-2019-sentence-bert}. Inferences with intermediate similarity levels (0.4 $\sim$ 0.7) to the correct answer are then randomly sampled, ensuring distractors remain contextually relevant yet challenging.

\begin{figure*}[t!]
\centering
  \includegraphics[width=\linewidth]{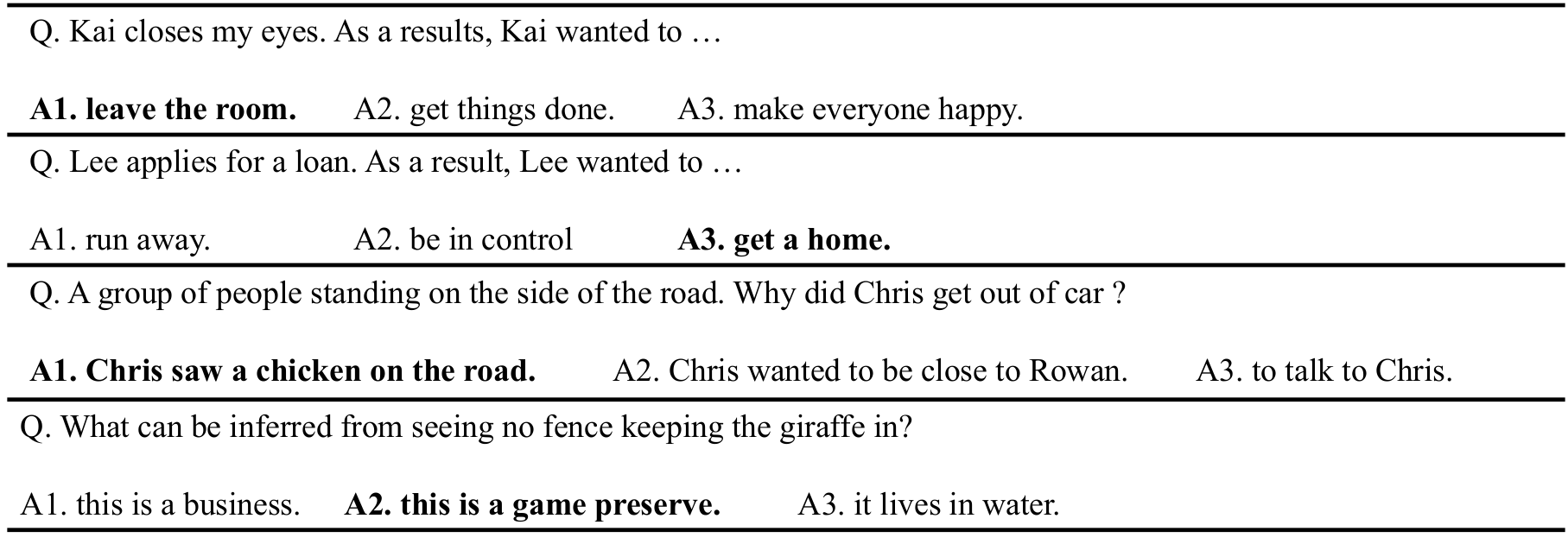}
    \caption{Examples of less plausible data filtered during the construction of Synthetic VQA$\boldsymbol{+}$. We measured the commonsense plausibility of (question, correct answer) pairs using the VERA model \citep{DBLP:conf/emnlp/0010WWS0H23-vera}. \textbf{Bold} indicates the correct answer.}
    \label{fig:filtered}
\end{figure*}

While augmenting the dataset improves its diversity, maintaining its correctness is equally critical. We observed that some samples in Synthetic VQA contain incorrect commonsense knowledge, which could mislead the model. To address this issue, we leverage the VERA model \citep{DBLP:conf/emnlp/0010WWS0H23-vera}, a commonsense correctness estimator, to filter out implausible samples. By calculating a plausibility score for each question-answer pair, we remove data falling below a predefined threshold (0.5). This two-step process—augmentation and filtering—ensures that Synthetic VQA$\boldsymbol{+}$ not only contains richer but also more reliable visual commonsense information.
The statistics for Synthetic VQA$\boldsymbol{+}$ are in Table \ref{tab:statistics_plus}.
We also provide examples of filtered data in Figure \ref{fig:filtered}.

\begin{table}[ht!]
\centering
% \resizebox{0.5\columnwidth}{!}{%
\scriptsize
\begin{tabular}{@{}l|cc|c@{}}
\toprule
 & Train & Dev & Total \\ \midrule
\# Images generated from AbsAT & 17,674 & 1,612 & 19,286 \\
\# QA pairs from AbsAT & 413,013 & 38,199 & 451,212 \\ \midrule
\# Images from VCR & 74,025 & 9,146 & 83,171 \\
\# QA pairs from VCR & 162,480 & 20,137 & 182,617 \\ \midrule
\# Images from Sherlock & 89,762 & 6,619 & 96,381 \\
\# QA pairs from Sherlock & 273,627 & 19,695 & 293,322 \\ \midrule
\# Total Images & 166,080 & 14,683 & 180,763 \\
\# Total QA pairs & 849,120 & 78,031 & 927,151 \\ \bottomrule
\end{tabular}%
% }
\caption{Statistic of Synthetic VQA+ dataset.}
\label{tab:statistics_plus}
\end{table}

\subsection{Pre-training \textsc{Imagine} on Synthetic VQA}
\label{subsec:injection}
Based on the Synthetic VQA dataset, we integrate commonsense knowledge into the models. Since \textsc{Imagine} involves two distinct modalities (i.e., text and image), we introduce two separate objectives to select the best answer choice: Language Modeling (LM) and Image-Text Matching (ITM). To obtain the LM scores, we calculate the masked language modeling loss for the Transformer encoder-based model, formulated as:
\begin{equation}
S_{LM}(T) = -\frac{1}{m} \sum_{t=1}^{m} \log P(w_t | ...w_{t-1}, w_{t+1}...).
\end{equation}
For the decoder-based model, we compute the auto-regressive language modeling loss, defined as:
\begin{equation}
S_{LM}(T) = -\frac{1}{m} \sum_{t=1}^{m} \log P(w_t | w_{1}...w_{t-1}),
\end{equation}
where $w_i$ denotes the $i$-th word, and $m$ is the number of tokens in the sequence $T$. 
To compute the ITM scores, we first contextualize the visual features based on the textual sequences. Let the visual features from the visual encoder $M_{I}$ be denoted as $V$, we derive the contextualized visual features as follows:
\begin{equation}
    C = \text{softmax}(\frac{\vec{T}V^{\top}}{\sqrt{d_v}})V,
\end{equation}
where $\vec{T}$ is the feature vector from the PLMs $M_{T}$. 
For the encoder-based model, we use the final hidden state of the $[\text{CLS}]$ token as the context vector, and for the decoder-based model, we use the hidden state of the last token as the context vector.\
% , and we use the final hidden state of $[\text{CLS}]$ token as the context vector. 
$d_v$ is the dimension of visual features. We then achieve the ITM scores by calculating the similarity between contextualized visual features and textual features as follows:
\begin{equation}
    S_{I}(T, V) = \text{sim}(\vec{T}, C),
\end{equation}
where $\text{sim}(\cdot)$ denotes the cosine similarity function. By combining two different scores, we produce the joint scores $S_{J}$ as follows:
\begin{equation}
    S_{J}(T, V) = \frac{1}{2}(S_{LM}(T) + S_{I}(T, V)),
\end{equation}
After calculating all scores $S^{(1)}, S^{(2)}, ..., S^{(n)}$ for $n$ answer candidates, we calculate the marginal ranking loss defined as:
\begin{equation}
    \mathcal{L}_{QA}(S) = \frac{1}{n}\sum_{i=1, i \neq y}^{n} \max(0, \eta -S^{(y)} + S^{(i)}),
\end{equation}
where $y$ indicates the index of the correct answer and $\eta$ is the pre-defined margin. The overall objectives are as follows:
\begin{equation}
    \mathcal{L} = \mathcal{L}_{QA}(S_{LM}) + \mathcal{L}_{QA}(S_I) + \mathcal{L}_{QA}(S_J).   
\end{equation}
However, we have empirically observed that the ITM objective prevents the model from learning the LM objective, which is essential for developing reasoning capabilities. To mitigate the conflict between these two objectives, we introduce two distinct adapters \citep{DBLP:conf/iclr/HeZMBN22-parallel_adapter}, LM adapter and ITM adapter. Each adapter is trained separately with a different focus. It is important to note that only the weights within these adapters are optimized during training; all other parameters remain frozen. By separating the parameters for objectives, we can effectively reduce conflicts between them.

\subsection{Inference from $\textsc{Imagine}$}
\label{subsec:inference}
For the inference, we use the same strategy to compute the LM and ITM scores after synthesizing the image based on the question.
Then, we assemble two scores to derive the model's prediction after obtaining the probability distribution through softmax.
\begin{equation}
    P(S) = \text{softmax}(S^{(1)}, S^{(2)}, ..., S^{(n)}),
\end{equation}
\begin{equation}
    P(A|Q) = (1-\lambda) \cdot P(S_M) + \lambda \cdot P(S_I),
\end{equation}
where $\lambda$ is an ensemble coefficient that controls the contributions between textual and visual features.

\subsection{Faster Inference via Image Retrieval} \label{faster}
Generating images for each input question offers rich contextual information but introduces significant computational overhead, especially when resources are limited. Moreover, generated images may occasionally misalign with the input question's context, leading to inaccuracies or ambiguities in representing real-world scenarios. Such limitations necessitate more efficient approaches, particularly when inference speed and precision are critical.

To overcome these challenges and significantly enhance inference speed, we propose a retrieval-based inference method within \textsc{Imagine}, which retrieves pre-existing images instead of generating new ones. We considered two retrieval approaches: a simple retrieval mechanism based on image-text similarity and an enhanced approach leveraging query-relevant textual knowledge obtained from ConceptNet \citep{DBLP:conf/aaai/SpeerCH17-conceptnet}. Ultimately, we selected the simpler retrieval method due to its computational efficiency and satisfactory performance. Additional discussions and evaluations of the ConceptNet-based retrieval approach are presented in Section \ref{dc:retrieval}.

For our chosen retrieval method, we first construct a comprehensive image database utilizing images from Synthetic VQA$\boldsymbol{+}$ and MSCOCO \citep{DBLP:conf/eccv/LinMBHPRDZ14}. Each input question and all database images are encoded using CLIP \citep{DBLP:conf/icml/RadfordKHRGASAM21-clip}. The top-1 most relevant image is then efficiently identified via cosine similarity, capitalizing on CLIP’s effectiveness in measuring high semantic similarity between image-text pairs. This streamlined retrieval approach ensures both speed and contextual alignment during inference.

\section{Experiments}
In this section, we evaluate the effectiveness of \textsc{Imagine} through extensive experiments and analysis. 
% Our investigation is guided by the following research questions:
% \begin{itemize}
%     \item Does \textsc{Imagine} achieve superior zero-shot performance across diverse reasoning benchmarks? (\S\ref{subsec: main}) 
    
%     \item Can \textsc{Imagine} effectively combine visual imagination with textual knowledge? (\S\ref{subsec: insight}, \S\ref{subsec: insight2})

%     \item What is the contribution of each component of \textsc{Imagine} to zero-shot commonsense reasoning? (\S\ref{subsec: trans})
    
% \end{itemize}

\subsection{Experimental Setup}
\paragraph{Dataset}
% 선정 이유
Following the previous works on zero-shot reasoning \citep{DBLP:conf/aaai/MaIFBNO21-ma, DBLP:conf/emnlp/YangYZWYC22-zlavi}, we evaluate our framework on commonsense reasoning tasks, science QA tasks, and natural language understanding tasks from GLUE benchmark to assess its generalizability\footnote{Evaluation results on NLU tasks are in Section \ref{app: nlu_tasks}.}.
Specifically, we evaluate each baseline on the five reasoning benchmarks, including Abductive NLI ($\alpha$NLI; \citet{DBLP:conf/iclr/BhagavatulaBMSH20-anli}), CommonsenseQA (CSQA; \citet{DBLP:conf/naacl/TalmorHLB19-csqa}), PhysicalIQA (PIQA; \citet{DBLP:conf/aaai/BiskZLGC20-piqa}), SocialIQA (SIQA; \citet{DBLP:conf/emnlp/SapRCBC19-socialiqa}), and Winogrande (WG; \citet{DBLP:conf/aaai/SakaguchiBBC20-winogrande}). These datasets vary significantly in format (e.g., natural language inference, QA, pronoun resolution) and required knowledge (e.g., social and physical knowledge for SIQA and PIQA, respectively), enabling a comprehensive evaluation of a wide spectrum of reasoning capabilities. For science QA tasks, we assess each baseline on the four benchmarks, including QA via Sentence Composition (QASC; \citet{DBLP:conf/aaai/KhotCGJS20-qasc}), Science Questions (SciQ; \citet{DBLP:conf/aclnut/WelblLG17-sciq}), and the AI2 Reasoning Challenge (ARC-Easy, ARC-Challenge; \citet{DBLP:journals/corr/abs-1803-05457-arc}). Given that science QA datasets often contain various types of reporting bias, such as color and shape biases, we selected these datasets to verify the efficacy of \textsc{Imagine} in mitigating reporting bias. 
Lastly, we evaluate our framework with a more general natural language understanding task to see if the integration of multi-modal inputs does not harm language comprehension ability.

\paragraph{Baselines}
We mainly compare \textsc{Imagine} with the following zero-shot commonsense reasoning frameworks:  MR \citep{DBLP:conf/aaai/MaIFBNO21-ma}, SMLM \citep{DBLP:conf/emnlp/BanerjeeB20-smlm}, Zero-shot Fusion \citep{DBLP:conf/naacl/KimKKAHY22}, CAR \citep{DBLP:conf/emnlp/WangF0XLSB23-car}, and the state-of-the-art framework, CANDLE \citep{DBLP:journals/corr/abs-2401-07286-candle}. To confirm the efficacy of training with machine imagination in \textsc{Imagine}, we also compare it with vision-language models \citet{DBLP:conf/nips/LiuLWL23a, DBLP:conf/nips/Dai0LTZW0FH23-instructblip} and Z-LaVI \citep{DBLP:conf/emnlp/YangYZWYC22-zlavi}, which leverages machine imagination but does not include the training process. Beyond the reasoning framework based on KBs, we evaluate the recent LLMs, which include LLaMA2$_{\text{13B}}$ \citep{DBLP:journals/corr/abs-2307-092880-llama2}, Mistral$_{\text{7B}}$ (v0.1) \citep{DBLP:journals/corr/abs-2310-06825-mistral}, OPT$_{\text{30B}}$ \citep{DBLP:journals/corr/abs-2205-01068-opt}, FLAN$_{\text{137B}}$ \citep{DBLP:conf/iclr/WeiBZGYLDDL22-flan}, and the GPT families (i.e., GPT-3, ChatGPT (\texttt{gpt-3.5-turbo)}, GPT-4). 

\paragraph{Backbones}
To verify the general applicability of \textsc{Imagine}, we apply our method to the both encoder and decoder models. Specifically, following the previous works, we utilize RoBERTa-Large \citep{DBLP:journals/corr/abs-1907-11692-roberta} and DeBERTa-v3-Large \citep{DBLP:conf/iclr/HeGC23-debertav3}. Each model has 362M and 443M parameters, respectively. As for the decoder model, we use GPT-2-Large that involves 792M parameters. 

\begin{table}[!t]
\centering
\scriptsize
\begin{tabular}{@{}l|ccc@{}}
\toprule
\textsc{Imagine} & GPT-2-L & RoBERTa-L & DeBERTa-v3-L \\ \midrule
\# Params. & 792M & 362M & 443M \\
\# Trainable Params. & 7.9M & 8.4M & 8.4M \\
Training Time & 70h & 30h & 80h \\  
Image Encoder & \multicolumn{3}{c}{CLIP-ViT-L/14} \\
Batch Size & \multicolumn{3}{c}{8, 16, \textbf{32}, 64} \\
Learning Rate & \multicolumn{3}{c}{7e-6, \textbf{1e-5}, 3e-5} \\
Epoch & \multicolumn{3}{c}{2} \\ \bottomrule

\end{tabular}
\caption{Detailed training settings for \textsc{Imagine}. \textbf{Bold} indicates the chosen hyperparameter.}
\label{tab:implementation_details}
\end{table}

\paragraph{Implementation details}
\label{app: details}
To construct the VQA pairs, we primarily use DALL-E 3-XL \citep{betker2023improving-dalle3}, a powerful image synthesis model. For generating images in the Synthetic VQA dataset, we first remove overly specific information, such as personal names, from the questions. Then, we generate images with a resolution of $512\times512$ using 50 inference steps. During the evaluation, we generate $512\times512$ images for each task based on the questions, maintaining the same number of inference steps.
We use the CLIP-Large \citep{DBLP:conf/icml/RadfordKHRGASAM21-clip} model to extract image features. Following prior work, we use two powerful PLMs as the backbone. We add Parallel Adapter \citep{DBLP:conf/iclr/HeZMBN22-parallel_adapter} with a reduction factor of 16 to each model and freeze all parameters except for the adapters. We follow the training settings of \citet{DBLP:conf/aaai/MaIFBNO21-ma} and \citet{DBLP:conf/emnlp/WangF0XLSB23-car} to train Transformer decoder-based and encoder-based model for the in-depth comparison. We report our results derived from the ensemble score using the optimal ensemble weight for each task.
All experiments are conducted using four NVIDIA A5000 GPUs. More details are presented in Table \ref{tab:implementation_details}.

\begin{table*}[t!]
\centering
\resizebox{\textwidth}{!}{%
\begin{tabular}{@{}lccccccc@{}}
\toprule
\multicolumn{1}{l|}{Method} & \multicolumn{1}{c|}{KB} & $\alpha$NLI & CSQA & PIQA & SIQA & \multicolumn{1}{c|}{WG} & Avg. \\ \midrule
\multicolumn{8}{l}{\textbf{Pre-trained Language Models}} \\
\multicolumn{1}{l|}{GPT-2-L \citep{radford2019language-gpt2}} & \multicolumn{1}{c|}{-} & 56.5 & 41.4 & 68.9 & 44.6 & \multicolumn{1}{c|}{53.2} & 52.9 \\
\multicolumn{1}{l|}{RoBERTa-L \citep{DBLP:journals/corr/abs-1907-11692-roberta}} & \multicolumn{1}{c|}{-} & 65.6 & 45.0 & 67.6 & 47.3 & \multicolumn{1}{c|}{57.5} & 56.6 \\
\multicolumn{1}{l|}{DeBERTa-v3-L \citep{DBLP:conf/iclr/HeGC23-debertav3}} & \multicolumn{1}{c|}{-} & 59.9 & 25.4 & 44.8 & 47.8 & \multicolumn{1}{c|}{50.3} & 45.6 \\
\multicolumn{1}{l|}{Self-talk \citep{DBLP:conf/emnlp/ShwartzWBBC20-selftalk}} & \multicolumn{1}{c|}{-} & - & 32.4 & 70.2 & 46.2 & \multicolumn{1}{c|}{54.7} & - \\
\multicolumn{1}{l|}{COMET-DynGen \citep{DBLP:conf/aaai/BosselutBC21-dyngen}} & \multicolumn{1}{c|}{AT} & - & - & - & 50.1 & \multicolumn{1}{c|}{-} & - \\
% \multicolumn{1}{l|}{SMLM \citep{DBLP:conf/emnlp/BanerjeeB20-smlm}} & \multicolumn{1}{c|}{*} & 65.3 & 38.8 & - & 48.5 & \multicolumn{1}{c|}{*} & - \\
\multicolumn{1}{l|}{GPT-2-L (MR; \citet{DBLP:conf/aaai/MaIFBNO21-ma})} & \multicolumn{1}{c|}{AT} & 59.2 & 48.0 & 67.5 & 53.6 & \multicolumn{1}{c|}{54.7} & 56.6 \\
\multicolumn{1}{l|}{RoBERTa-L (MR; \citet{DBLP:conf/aaai/MaIFBNO21-ma})} & \multicolumn{1}{c|}{AT} & 70.8 & 64.2 & 72.1 & 63.1 & \multicolumn{1}{c|}{59.6} & 66.0 \\
\multicolumn{1}{l|}{DeBERTa-v3-L (MR; \citet{DBLP:conf/aaai/MaIFBNO21-ma})} & \multicolumn{1}{c|}{AT} & 76.0 & 67.0 & 78.0 & 62.1 & \multicolumn{1}{c|}{76.0} & 71.8 \\
\multicolumn{1}{l|}{MICO \citep{DBLP:conf/emnlp/SuWFZSZ22-mico}} & \multicolumn{1}{c|}{AT} & - & 44.2 & - & 56.0 & \multicolumn{1}{c|}{-} & - \\
\multicolumn{1}{l|}{Zero-shot Fusion \citep{DBLP:conf/naacl/KimKKAHY22}} & \multicolumn{1}{c|}{AT, CN, WD, WN} & 72.5 & 68.2 & 72.9 & 66.6 & \multicolumn{1}{c|}{60.8} & 68.2 \\
\multicolumn{1}{l|}{Multi-hop Knowledge Injection \citep{DBLP:journals/corr/abs-2305-05936-multi-hop}} & \multicolumn{1}{c|}{AT, CN, WD, WN} & 72.5 & 71.0 & 73.1 & - & \multicolumn{1}{c|}{61.0} & - \\
\multicolumn{1}{l|}{CAR-GPT-2-L \citep{DBLP:conf/emnlp/WangF0XLSB23-car}} & \multicolumn{1}{c|}{AbsAT} & 61.7 & 50.0 & 68.2 & 52.3 & \multicolumn{1}{c|}{55.2} & 57.5 \\
\multicolumn{1}{l|}{CAR-RoBERTa-L \citep{DBLP:conf/emnlp/WangF0XLSB23-car}} & \multicolumn{1}{c|}{AbsAT} & 72.7 & 66.3 & 73.2 & 64.0 & \multicolumn{1}{c|}{62.0} & 67.6 \\
\multicolumn{1}{l|}{CAR-DeBERTa-v3-L \citep{DBLP:conf/emnlp/WangF0XLSB23-car}} & \multicolumn{1}{c|}{AbsAT} & 79.6 & 69.3 & 78.6 & 64.0 & \multicolumn{1}{c|}{78.2} & 73.9 \\
\multicolumn{1}{l|}{CANDLE-DeBERTa-v3-L \citep{DBLP:journals/corr/abs-2401-07286-candle}} & \multicolumn{1}{c|}{CANDLE} & 81.2 & 69.9 & 80.3 & 65.9 & \multicolumn{1}{c|}{78.3} & 75.1 \\ \midrule
\multicolumn{8}{l}{\textbf{Large Language Models}} \\
\multicolumn{1}{l|}{GPT-3.5 (\texttt{text-davinci-003})} & \multicolumn{1}{c|}{-} & 61.8 & 68.9 & 67.8 & 68.0 & \multicolumn{1}{c|}{60.7} & 65.4 \\
\multicolumn{1}{l|}{ChatGPT (\texttt{gpt-3.5-turbo})} & \multicolumn{1}{c|}{-} & 73.2 & 75.7 & 81.7 & \textbf{69.7} & \multicolumn{1}{c|}{64.1} & 72.9 \\
\multicolumn{1}{l|}{GPT-4 (\texttt{gpt-4})} & \multicolumn{1}{c|}{-} & 75.0 & 43.0 & 73.0 & 57.0 & \multicolumn{1}{c|}{77.0} & 65.0 \\
\multicolumn{1}{l|}{LLAMA2-13B \citep{DBLP:journals/corr/abs-2307-092880-llama2}} & \multicolumn{1}{c|}{-} & 55.9 & 67.3 & 80.2 & 50.3 & \multicolumn{1}{c|}{72.8} & 65.3 \\
\multicolumn{1}{l|}{Mistral-v0.1-7B \citep{DBLP:journals/corr/abs-2310-06825-mistral}} & \multicolumn{1}{c|}{-} & 51.0 & 59.6 & \textbf{83.0} & 42.9 & \multicolumn{1}{c|}{75.3} & 62.4 \\
\multicolumn{1}{l|}{VERA-T5-xxl \citep{DBLP:conf/emnlp/0010WWS0H23-vera}} & \multicolumn{1}{c|}{AT} & 71.2 & 61.7 & 76.4 & 58.2 & \multicolumn{1}{c|}{67.2} & 66.9 \\
\multicolumn{1}{l|}{VERA-T5-xxl \citep{DBLP:conf/emnlp/0010WWS0H23-vera}} & \multicolumn{1}{c|}{AbsAT} & 73.2 & 63.0 & 77.2 & 58.1 & \multicolumn{1}{c|}{68.1} & 68.0 \\
\multicolumn{1}{l|}{CANDLE-VERA-T5-xxl \citep{DBLP:journals/corr/abs-2401-07286-candle}} & \multicolumn{1}{c|}{CANDLE} & 73.8 & 64.7 & 77.6 & 59.4 & \multicolumn{1}{c|}{71.3} & 69.4 \\ \midrule
\multicolumn{8}{l}{\textbf{Vision-Language Models}} \\
\multicolumn{1}{l|}{LLaVA-1.5-7B \citep{DBLP:conf/nips/LiuLWL23a}} & \multicolumn{1}{c|}{-} & 55.2 & 29.4 & 64.2 & 34.8 & \multicolumn{1}{c|}{54.5} & 47.6 \\
\multicolumn{1}{l|}{InstructBLIP-Vicuna-7B \citep{DBLP:conf/nips/Dai0LTZW0FH23-instructblip}} & \multicolumn{1}{c|}{-} & 54.8 & 40.5 & 66.0 & 42.1 & \multicolumn{1}{c|}{59.6} & 52.6 \\  \midrule
\multicolumn{8}{l}{\textbf{Ours}} \\
\rowcolor{lightblue}
\multicolumn{1}{l|}{\textsc{Imagine}-GPT-2-L} & \multicolumn{1}{c|}{Synthetic VQA} & 61.5 & 63.9 & 58.9 & 53.0 & \multicolumn{1}{c|}{55.2} & 58.5 \\
\rowcolor{lightblue}
\multicolumn{1}{l|}{\textsc{Imagine}-RoBERTa-L} & \multicolumn{1}{c|}{Synthetic VQA} & 74.7 & 67.5 & 72.3 & 64.3 & \multicolumn{1}{c|}{61.2} & 68.0 \\
\rowcolor{lightblue}
\multicolumn{1}{l|}{\textsc{Imagine}-DeBERTa-v3-L} & \multicolumn{1}{c|}{Synthetic VQA} & 82.2 & 74.0 & 80.7 & 66.3 & \multicolumn{1}{c|}{76.7} & 76.0 \\ 
\rowcolor{lightblue}
\multicolumn{1}{l|}{\textsc{Imagine}-DeBERTa-v3-L} & \multicolumn{1}{c|}{Synthetic VQA$\boldsymbol{+}$} & \textbf{83.4} & \textbf{76.3} & \underline{81.4} & \underline{69.0} & \multicolumn{1}{c|}{\textbf{79.3}} & \textbf{77.9}  \\ 
\rowcolor{lightblue}
\multicolumn{1}{l|}{\textsc{Imagine}-DeBERTa-v3-L (Retrieval)} & \multicolumn{1}{c|}{Synthetic VQA$\boldsymbol{+}$} & \underline{83.3} & \underline{76.2} & 81.3 & 69.0 & \multicolumn{1}{c|}{\underline{79.2}} & \underline{77.8}  \\ \midrule
\multicolumn{8}{l}{\textbf{Supervised \& Human}} \\
\multicolumn{1}{l|}{RoBERTa-L (Supervised)} & \multicolumn{1}{c|}{-} & 85.6 & 78.5 & 79.2 & 76.6 & \multicolumn{1}{c|}{79.3} & 79.8 \\
\multicolumn{1}{l|}{DeBERTa-v3-L (Supervised)} & \multicolumn{1}{c|}{-} & 89.0 & 82.1 & 84.5 & 80.1 & \multicolumn{1}{c|}{84.1} & 84.0 \\ 
\multicolumn{1}{l|}{Human} & \multicolumn{1}{c|}{-} & 91.4 & 88.9 & 94.9 & 86.9 & \multicolumn{1}{c|}{94.1} & 91.2 \\ \bottomrule
\end{tabular}%
}
\caption{Zero-shot evaluation results on five commonsense reasoning tasks (Accuracy \%). \textbf{Bold} and 
\underline{Underline} indicate the best and second-best results, respectively. AT, CN, WD, WN, and AbsAT refer to ATOMIC, ConcetNet, WikiData, WordNet, and AbstractATOMIC. The results of the large language models including GPT series are taken from \citet{DBLP:journals/corr/abs-2401-07286-candle}. (Retrieval) refers to the model inference using retrieved images instead of generating them, as described in Section \ref{subsec:inference}.}
% SMLM (*) used different KBs for the different benchmarks.}
\label{tab:csqas}
\end{table*}

\begin{table}[t!]
% \small
\centering
\resizebox{\textwidth}{!}{%
\begin{tabular}{@{}l|c|cccc@{}}
\toprule
Method & KB & QASC & SciQ & ARC-E & ARC-C \\ \midrule
SMLM \citep{DBLP:conf/emnlp/BanerjeeB20-smlm} & * & 26.6 & - & 33.4 & 28.4 \\
GPT-Neo-2.7B \citep{gpt-neo} & - & 29.6 & 64.0 & 49.6 & 31.8 \\
GPT-J-6B \citep{gpt-j} & - & 36.3 & 73.2 & 44.1 & 34.8 \\
CAR-RoBERTa-L \citep{DBLP:conf/emnlp/WangF0XLSB23-car} & AbsAT & 56.7 & 60.7 & 57.0 & 36.5 \\
CAR-DeBERTa-v3-L \citep{DBLP:conf/emnlp/WangF0XLSB23-car} & AbsAT & 70.0 & 76.9 & 75.3 & 53.2 \\
OPT-30B \citep{DBLP:journals/corr/abs-2205-01068-opt} & - & 39.7 & 72.7 & 58.2 & 34.8 \\
FLAN-137B \citep{DBLP:conf/iclr/WeiBZGYLDDL22-flan} & - & - & - & \textbf{79.5} & \textbf{61.7} \\ \midrule
Z-LaVI (RoBERTa-L) \citep{DBLP:conf/emnlp/YangYZWYC22-zlavi} & - & 27.2 & 51.3 & 51.8 & 33.4 \\
Z-LaVI (BART-L)\citep{DBLP:conf/emnlp/YangYZWYC22-zlavi} & - & 27.3 & 51.0 & 56.1 & 36.5 \\
Z-LaVI (OPT-30B)\citep{DBLP:conf/emnlp/YangYZWYC22-zlavi} & - & 42.1 & 74.0 & 59.5 & 34.1 \\
\rowcolor{lightblue} \textsc{Imagine}-GPT-2-L & Synthetic VQA & 46.5 & 58.4 & 55.1 & 35.1 \\
\rowcolor{lightblue} \textsc{Imagine}-RoBERTa-L & Synthetic VQA & 57.1 & 63.7 & 57.9 & 39.1 \\
\rowcolor{lightblue} \textsc{Imagine}-DeBERTa-v3-L & Synthetic VQA & 72.4 & 78.9 & 76.0 & 56.2 \\
\rowcolor{lightblue} \textsc{Imagine}-DeBERTa-v3-L & Synthetic VQA$\boldsymbol{+}$ & \textbf{73.5} & \underline{80.5} & \underline{79.1} & 59.2 \\
\rowcolor{lightblue} \textsc{Imagine}-DeBERTa-v3-L (Retrieval) & Synthetic VQA$\boldsymbol{+}$ & \underline{73.5} & \textbf{80.7} & 78.9 & \underline{59.5}  \\
\bottomrule
\end{tabular}%
}
\caption{Zero-shot evaluation results on four science question-answering tasks (Accuracy \%). \textbf{Bold} and \underline{Underline} indicate the best and second-best results, respectively. SMLM (*) used different KBs for the different benchmarks. (Retrieval) refers to the model inference using retrieved images instead of generating them, as described in Section \ref{subsec:inference}.}
\label{tab:sciences}
\end{table}

\subsection{Experimental Results}
\label{subsec: main}

Tables \ref{tab:csqas}, and \ref{tab:sciences} show the results for the commonsense reasoning tasks and the science question-answering tasks. Models based on \textsc{Imagine} reveal either superior or competitive performance on overall reasoning tasks. This demonstrates the effectiveness of \textsc{Imagine} and highlights the benefit of leveraging machine imagination for reasoning. 

\paragraph{Comparison with Zero-shot Reasoners}
When compared to zero-shot commonsense reasoning frameworks in commonsense reasoning tasks (Table  \ref{tab:csqas}), \textsc{Imagine}-DeBERTa-v3-L model surpasses the previous state-of-the-art CANDLE-DeBERTa-v3-L by 2.8\%p on average, and specifically by 6.4\%p on the CSQA. This suggests that Synthetic VQA significantly enhances generalization performance in zero-shot commonsense reasoning.

\paragraph{Comparison with LLMs}
Comparison results with LLMs also show that \textsc{Imagine} outperforms recent LLMs, including ChatGPT and GPT-4 \citep{DBLP:journals/corr/abs-2303-08774-gpt4}. These findings highlight the superior efficiency and effectiveness of \textsc{Imagine}'s multimodal approach, underscoring its ability to integrate visual and textual information to surpass the capabilities of even the most advanced LLMs.

\paragraph{Comparison with VLMs}
We also provide results from recent powerful vision-language (VL) models (LLaVA-1.5 \citep{DBLP:conf/nips/LiuLWL23a}, InstructBLIP with Vicuna-7B \citep{DBLP:conf/nips/Dai0LTZW0FH23-instructblip}) by feeding the generated images. The results of vision-language models in Table \ref{tab:csqas} indicate that these VL models struggle to reason accurately about commonsense questions. We suspect that this issue arises from VL models’ tendency to focus on the image scene more than on textual inputs, as they are primarily trained to ground textual information into the image scene.
The datasets we experiment with prioritize linguistic ability over vision-language grounding and require reasoning rooted in commonsense knowledge. As a result, VL models that are more focused on visual understanding may underperform in zero-shot commonsense reasoning tasks, where linguistic reasoning ability is crucial.

\paragraph{Generalizability of \textsc{Imagine}}
\textsc{Imagine} also proves effective for science QA tasks (Table \ref{tab:sciences}). Compared to the models with KBs and larger models, \textsc{Imagine} presents better or competitive reasoning performance. These results confirm the effectiveness of the machine imagination capabilities on science-related contexts. We also highlight the comparison results with Z-LaVI \citep{DBLP:conf/emnlp/YangYZWYC22-zlavi} that leverages imagination similar to ours. \textsc{Imagine} outperforms this method by a significant margin (18.5\%p on average), underscoring the importance of the pre-training phase in effectively utilizing machine imagination.

\begin{figure*}[t!]
\centering
  \includegraphics[width=\columnwidth]{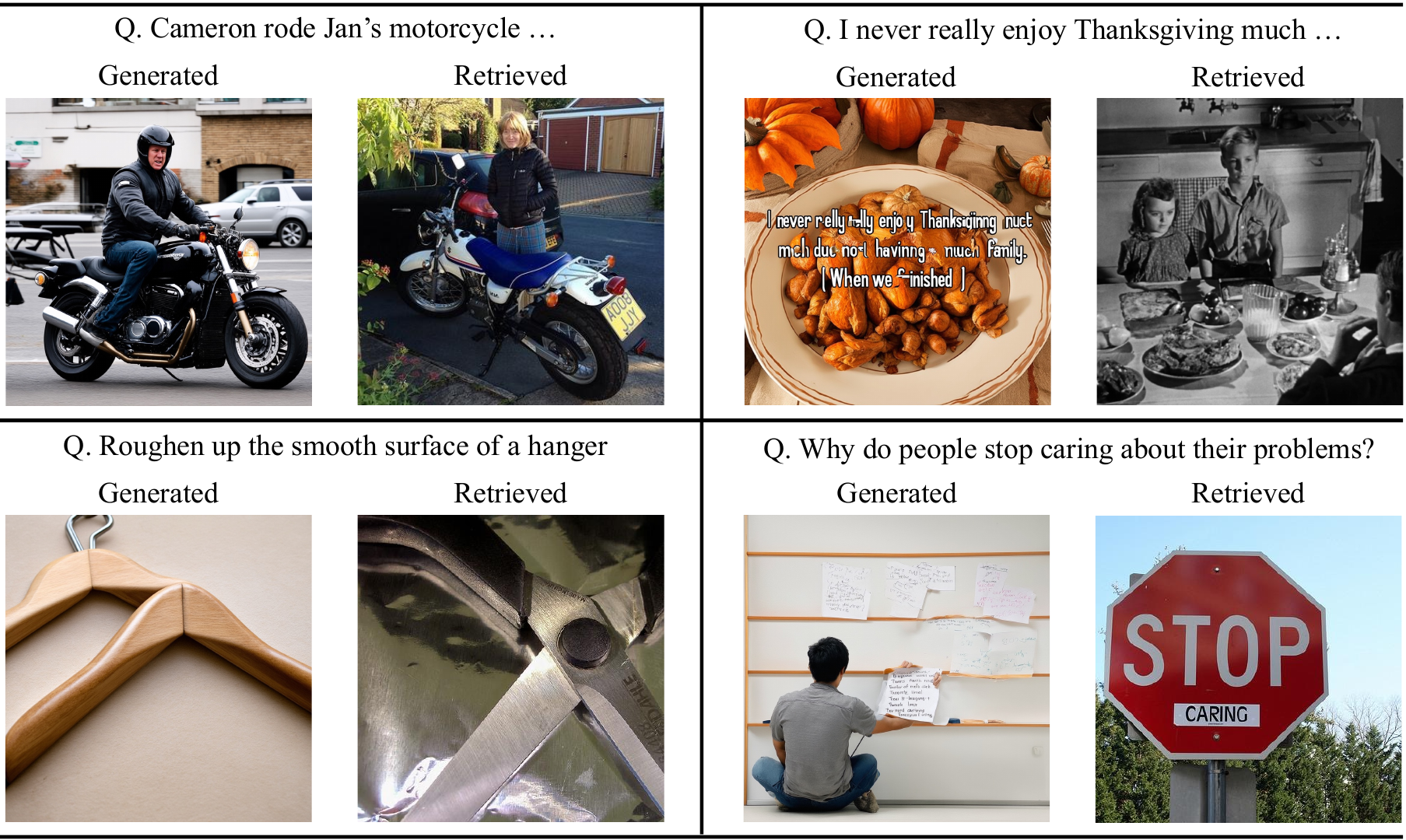}
    \caption{Examples of generated and retrieved images based on the input question. The first row shows cases where retrieved images are helpful for the inference, while the second row shows cases where retrieved images are not helpful.}
    \label{fig:retrieved_examples}
\end{figure*}

\paragraph{Impact of Synthetic VQA$\boldsymbol{+}$}
The effectiveness of Synthetic VQA$\boldsymbol{+}$ in enhancing zero-shot commonsense reasoning and science question-answering tasks is supported by our experiments. As an extended version of the original Synthetic VQA, Synthetic VQA$\boldsymbol{+}$ incorporates a broader and more diverse set of visual commonsense examples while filtering out implausible statements to ensure higher data quality. These refinements lead to an average performance improvement of 1.9\% over Synthetic VQA on commonsense reasoning tasks (Table \ref{tab:csqas}). It underscores the importance of data quality and diversity in achieving robust zero-shot reasoning. By aligning linguistic and visual contexts more effectively, Synthetic VQA$\boldsymbol{+}$ helps \textsc{Imagine} generalize better to unseen scenarios. 
% This improvement demonstrates the value of refining multimodal datasets to support zero-shot reasoning frameworks.

\paragraph{Results from Retrieval-based Inference} \label{retrieval_results}
Tables \ref{tab:csqas} and \ref{tab:sciences} demonstrate that the retrieval-based inference method achieves competitive performance compared to the generation-based approach. For instance, retrieval-based inference attains the highest accuracy (80.7\%) on the SciQ dataset, while maintaining reasonable performance across other tasks. Although the generation-based approach provides slightly higher reasoning accuracy, it requires approximately 21.5 seconds per inference. In contrast, retrieval-based inference completes within 1 second, making it significantly more efficient, especially in resource-constrained environments. A comparison of the retrieved and generated images is shown in Figure \ref{fig:retrieved_examples}.

\section{Discussions}

\begin{table}[t!]
\scriptsize
\centering
\begin{tabular}{@{}l|ccccc|c@{}}
\toprule
 & $\alpha$NLI & CSQA & PIQA & SIQA & WQ & Avg. \\ \midrule
DeBERTa-v3-L & 59.9 & 25.4 & 44.8 & 47.8 & 50.3 & 45.6 \\ \midrule
+ AbstractATOMIC  & 79.7 & 72.6 & 79.0 & 68.0 & 74.3 & 74.7 \\
+ VCR & 82.2 & 74.0 & 80.7 & 66.3 & 76.7 & 76.0 \\ %\midrule
+ Sherlock & 83.0 & 72.9 & 81.2 & 67.6 & 78.1 & 76.6 \\
+ Filtering & \textbf{83.4} & \textbf{76.3} & \textbf{81.4} & \textbf{69.0} & \textbf{79.2} & \textbf{78.8} \\ \bottomrule
\end{tabular}
\caption{Ablation results on Synthetic VQA$\boldsymbol{+}$. \textbf{Bold} indicates the best results. Filtering refers to the removal of implausible question-answer pairs using the VERA model\citep{DBLP:conf/emnlp/0010WWS0H23-vera} (Section \ref{subsec:vqa}).}
\label{tab:kb_ablation}
\end{table} 

\subsection{Contributions of Synthetic VQA$\boldsymbol{+}$}\label{subsec: insight2}
To confirm the effectiveness of each component in Synthetic VQA$\boldsymbol{+}$, we evaluate the contribution of each knowledge base and the filtering process in Section \ref{subsec:vqa}. 
The results in Table \ref{tab:kb_ablation} highlight the significant contributions of each component to the overall performance of Synthetic VQA$\boldsymbol{+}$. Incorporating knowledge bases such as AbstractATOMIC, VCR, and Sherlock leads to substantial improvements, with average accuracy rising from 45.6 (baseline) to 76.6 when these resources are combined. Furthermore, the filtering process, which removes implausible commonsense statements using the VERA model, achieves the highest performance, pushing the average accuracy to 78.8. This demonstrates the critical role of high-quality knowledge integration and effective filtering in enhancing zero-shot commonsense reasoning capabilities across diverse tasks.

\begin{table*}[t!]
% \small
\scriptsize
\centering
\resizebox{\columnwidth}{!}{%
\begin{tabular}{@{}l|cc|ccccccccccc@{}}
\toprule
 & AOKVQA & VCR & Synthetic VQA & Synthetic VQA$\boldsymbol{+}$ & $\alpha$NLI & CSQA & PIQA & SIQA & WG  \\ \midrule
Relevance & 23.81 & 21.26 & 23.59 & 24.85 & 30.26 & 29.38 & 30.80 & 29.92 & 29.26  \\ \bottomrule
\end{tabular}%
}
\caption{Image-text relevance evaluation using CLIP model.}
\label{tab:relevance}
\end{table*}

\subsection{Validation of Synthetic Dataset Quality}
We evaluate the quality of the dataset by measuring the relevance between the question text and the machine-generated images. For this purpose, inspired by the CLIP scores \citep{DBLP:conf/emnlp/HesselHFBC21}, we measure the relevance score between images and text using the CLIP model. A higher relevance score between the two modalities indicates that the image effectively captures the content of the text. As shown in Figure \ref{fig:qual}, images that are highly relevant to the questions can help to reason about the question.

First, we measure the relevance of datasets containing two multi-modal reasoning datasets of real images (A-OKVQA \citep{DBLP:conf/eccv/SchwenkKCMM22-aokvqa}, VCR \citep{DBLP:conf/cvpr/ZellersBFC19-vcr}) to establish a baseline. Then we compare these scores with those of the Synthetic VQA$\boldsymbol{+}$ and the synthetic pairs of all evaluation datasets to determine the data quality. The results in Table \ref{tab:relevance} show that most datasets exhibit similar or even higher relevance scores compared to the datasets containing real images (A-OKVQA, VCR). In particular, for two Synthetic VQA datasets, we evaluate only the dataset extracted from AbstractATOMIC, which contains only machine-generated images, and found that it has relevance scores closest to those of the real-image datasets. This demonstrates that our synthetic dataset has a quality comparable to that of the real VL dataset.

\begin{table*}[t!]
\centering
\resizebox{\textwidth}{!}{%
\begin{tabular}{@{}lccccccc@{}}
\toprule
\multicolumn{1}{l|}{Imagine-DeBERTa-v3-L} & \multicolumn{1}{c|}{KB} & $\alpha$NLI & CSQA & PIQA & SIQA & \multicolumn{1}{c|}{WG} & Avg. \\ \midrule
\multicolumn{1}{l|}{CLIP Retrieval} & \multicolumn{1}{c|}{Synthetic VQA$\boldsymbol{+}$} & 83.3 & \textbf{76.2} & \textbf{81.3} & 69.0 & \multicolumn{1}{c|}{\textbf{79.2}} & \textbf{77.8}  \\ 
\multicolumn{1}{l|}{Concept-aware Retrieval} & \multicolumn{1}{c|}{Synthetic VQA$\boldsymbol{+}$} & \textbf{83.8} & 75.5 & 81.2 & \textbf{70.1} & \multicolumn{1}{c|}{77.6} & 77.6  \\ \bottomrule
\end{tabular}%
}
\caption{Comparison of image-text retrieval approaches. CLIP retrieval is performed based solely on CLIP embedding similarities, whereas concept-aware retrieval enhances this process by integrating query-relevant knowledge from ConceptNet prior to similarity computation.}
\label{tab:retrieval}
\end{table*}

\subsection{Analysis on Image Retrieval} \label{dc:retrieval}
Table \ref{tab:retrieval} shows that among the two retrieval approaches we considered in Section \ref{faster}, the simpler CLIP embedding-based retrieval consistently achieves higher average accuracy (77.8\%) than Concept-aware retrieval (77.6\%). This performance gap is likely due to Concept-aware retrieval occasionally introducing irrelevant or less relevant knowledge from ConceptNet.
These findings are particularly pronounced in datasets like Winogrande, which specifically require reasoning based on subtle contextual clues rather than general commonsense knowledge alone. Consequently, our results highlight the need for adaptive, context-aware image retrieval techniques tailored to the unique characteristics of different reasoning tasks.

\subsection{Impact of Imagination on Model Inference}\label{subsec: insight}
\begin{figure*}[t!]
\centering
  \includegraphics[width=\linewidth]{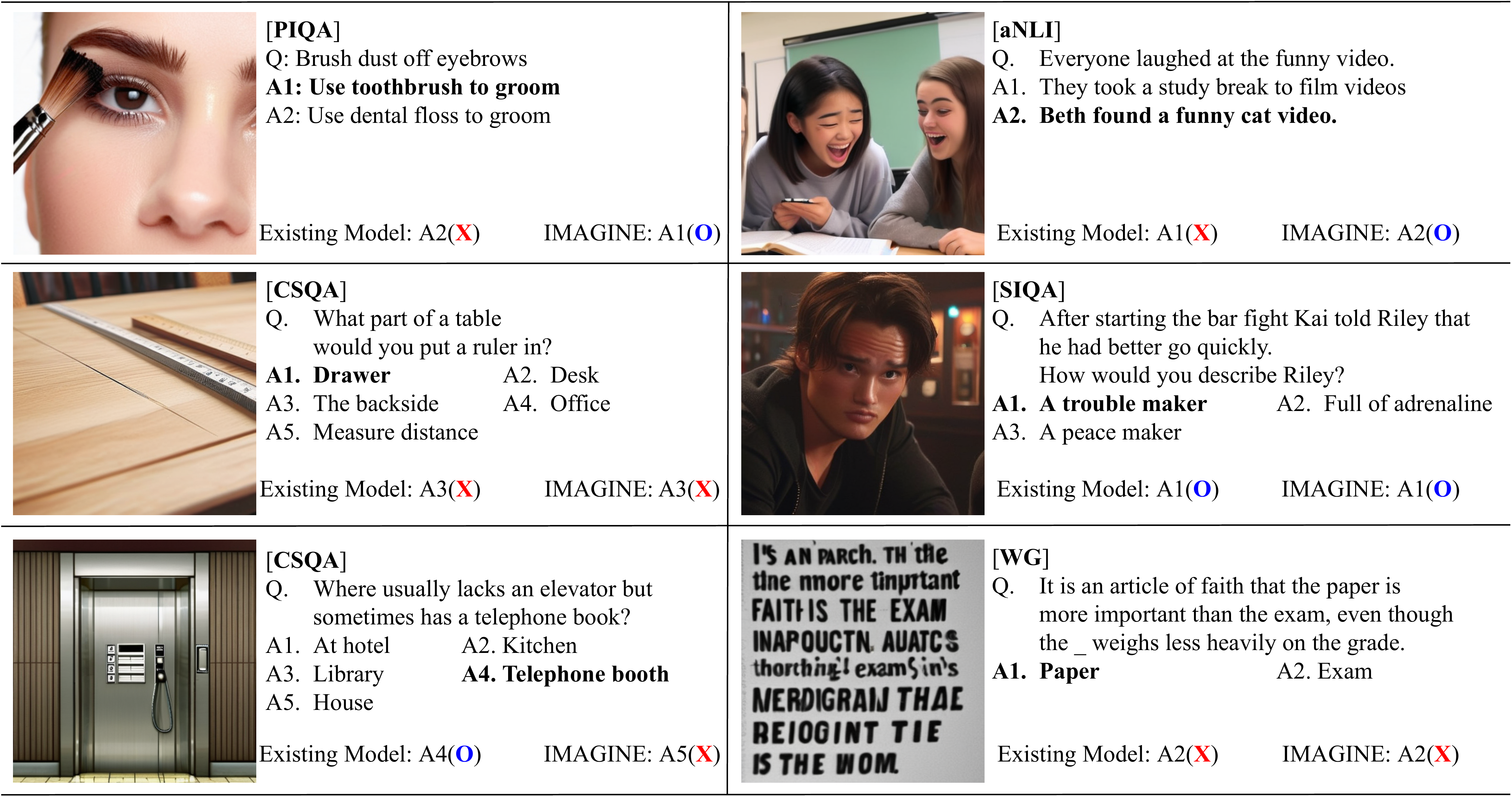}
    \caption{Comparison of model predictions and the correctness from \textsc{Imagine} and the existing model \citep{DBLP:conf/emnlp/WangF0XLSB23-car} on five commonsense reasoning tasks.}
    \label{fig:qual}
\end{figure*}

We analyze the inference results from the text-based model, CAR \citep{DBLP:conf/emnlp/WangF0XLSB23-car}, and \textsc{Imagine} to confirm the impact of machine imagination on the model inference. The results are shown in Figure \ref{fig:qual}. We draw three major findings regarding the impact of imagination: (i) When the text contains limited commonsense knowledge, imagination indeed helps the model to correctly infer the answer (First row in the Figure), i.e., positive impact on predictions (ii) When the generated images only partially capture the context of the text query, imagination does not affect the inference results (Second row in the Figure). (iii) When images deviate from the real world, imagination can lead to incorrect inferences (Third row in the Figure). Specifically, we empirically observe that longer text queries often result in such cases. 

To further assess how often images negatively impact model inference, we calculate the ratio of helpful imagination (i.e., imagination leading to correct reasoning) to harmful imagination (i.e., imagination leading to incorrect reasoning) across different commonsense reasoning benchmarks (Table \ref{tab: is_helpful_1} and \ref{tab: is_helpful_2}). Our analysis shows that helpful imagination contributes more than harmful imagination, suggesting that imagination generally has a positive impact. However, we also observe that in certain cases, misaligned imagination can lead to reasoning errors.

These results suggest that incorporating a text-to-image model with better alignment capabilities could potentially mitigate the negative impacts of imagination. We provide more examples with the visualization of model attention in \ref{app: Attention}.

\begin{table}[t!]
\centering
\begin{minipage}{0.48\textwidth}
\small
\centering
\resizebox{\textwidth}{!}{%
\begin{tabular}{@{}l|ccccc@{}}
\toprule
Imagine & aNLI & CSQA & PIQA & SIQA & WG \\ \midrule
Helpful (\%) & 30.2 & 41.1 & 26.9 & 36.2 & 11.8 \\
Harmful (\%) & 8.0 & 5.6 & 8.1 & 9.2 & 2.8 \\ \bottomrule
\end{tabular}%
}
\caption{Reliance on machine-generated images using \textsc{Imagine}-DeBERTa-v3-L}
\label{tab: is_helpful_1}
\end{minipage}
\hfill
\begin{minipage}{0.48\textwidth}
\small
\centering
\resizebox{\textwidth}{!}{%
\begin{tabular}{@{}l|ccccc@{}}
\toprule
Imagine & aNLI & CSQA & PIQA & SIQA & WG \\ \midrule
Helpful (\%) & 2.5 & 4.7 & 3.4 & 2.6 & 0.6 \\
Harmful (\%) & 1.7 & 3.7 & 2.7 & 1.3 & 0.5 \\ \bottomrule
\end{tabular}%
}
\caption{Reliance on machine-generated images using \textsc{Imagine}-RoBERTa-L}
\label{tab: is_helpful_2}
\end{minipage}
\end{table}

\subsection{Versatility of \textsc{Imagine}}
\label{app: nlu_tasks}
To assess the versatility of \textsc{Imagine}, we evaluate its zero-shot performance on general natural language understanding (NLU) tasks from the GLUE benchmark, including sentiment analysis (SST-2), natural language inference (RTE), and multi-genre inference (MNLI) \citep{DBLP:conf/iclr/WangSMHLB19}.

As shown in Table~\ref{tab: nlu_tasks}, \textsc{Imagine}-DeBERTa-v3-L consistently outperforms the base DeBERTa-v3-L model across all tasks, achieving particularly large gains on SST-2 (81.1 vs. 49.1) and MNLI (33.3 vs. 32.0). Compared to CAR-DeBERTa-v3-L, a model that incorporates external knowledge without using visual imagination, \textsc{Imagine} still shows competitive or superior results, especially on SST-2 and MNLI. This highlights the effectiveness of incorporating visual imagination in enhancing language understanding. In particular, the substantial improvement on SST-2 suggests that imagined visual contexts can provide strong complementary cues for tasks involving sentiment or emotionally grounded interpretation. These results demonstrate that \textsc{Imagine} offers a generalizable framework that enhances zero-shot performance by enriching textual representations with visual signals.

\begin{table}[t!]
\scriptsize
\centering
% \resizebox{\columnwidth}{!}{%
\begin{tabular}{@{}lccc@{}}
\toprule
Method & SST-2 & RTE & MNLI\\ \midrule
DeBERTa-v3-L & 49.1 & 50.5 & 32.0 \\
CAR-DeBERTa-v3-L & 56.2 & \textbf{54.9} & 32.0 \\
\textsc{Imagine}-DeBERTa-v3-L & \textbf{81.1} & 53.8 & \textbf{33.3} \\ \bottomrule
\end{tabular}%
% }
\caption{Zero-shot evaluation results on natural language understanding tasks.}
\label{tab: nlu_tasks}
\end{table}

\section{Conclusion}
In this paper, we propose \textsc{Imagine}, a novel zero-shot commonsense reasoning framework that leverages visual signals to mitigate reporting bias in textual inputs. We achieve this by coupling a pre-trained language model with a conditional image generator and image retriever, empowering the model with the capability of machine imagination. To steer \textsc{Imagine} in effectively utilizing visual information, we have created a large-scale Synthetic VQA$\boldsymbol{+}$ dataset and optimized the model to use both textual and visual information. Comprehensive evaluation results on various reasoning tasks show that \textsc{Imagine} establishes new state-of-the-art results on zero-shot commonsense reasoning tasks compared to strong baselines (including large language models), demonstrating the efficacy of machine imagination. 

% \section{Acknowledgements}
% This work was supported by the National Research Foundation of Korea (NRF) grant funded by the Korea government (MSIT) (No.RS-2024-00415812 and No.RS-2025-00517221) and Institute of Information \& communications Technology Planning \& Evaluation (IITP) grant funded by the Korea government (MSIT) (No.RS-2024-00439328, Karma: Towards Knowledge Augmentation for Complex Reasoning (SW Starlab), 
% No.RS-2024-00457882, AI Research Hub Project, and No.RS-2019-II190079, Artificial Intelligence Graduate School Program (Korea University)).  

\bibliographystyle{elsarticle-harv} 
\bibliography{eswa_base}

\newpage
\appendix

\begin{center}
\large
\textbf{Appendix}    
\end{center}

\section{Impact of Image Quality}
\label{app: c}
We aim to observe the changes in inference performance based on image quality by generating images of various qualities using three different methods. First, similar to our main experiment, we utilize the questions from the evaluation dataset to generate images with a resolution of $512\times512$ using both DALL-E 3-XL and the Latent Diffusion Model (LDM; \citet{DBLP:conf/cvpr/RombachBLEO22-stable}), which has relatively lower image synthesis capabilities. Additionally, we generate images with a resolution of $384\times384$ using DALL-E 3-XL, following the same method used for creating the Synthetic VQA dataset.

\begin{figure}[h!]
\centering
  \includegraphics[width=0.7\linewidth]{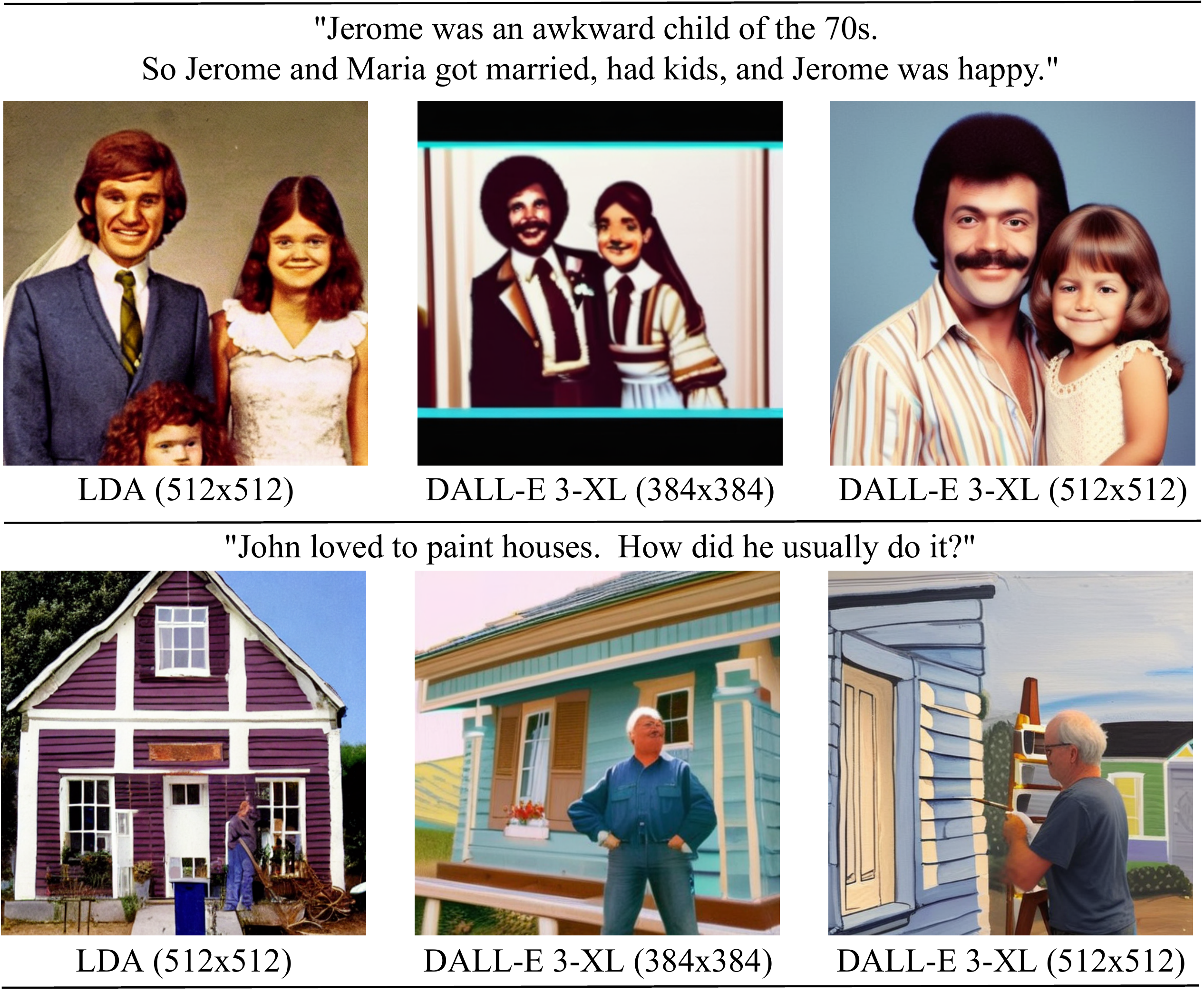}
    \caption{Comparison of generated images. The sentences are the queries used to generate the images.}
    \label{fig:img_quality}
\end{figure}

The results in Table \ref{tab:img_quality} show that the \textsc{Imagine} with the LDM model performs the worst, indicating that utilizing a less effective image synthesis model can degrade overall performance. However, all models benefit from incorporating various resolutions of images. As seen in Figure \ref{fig:img_quality}, this is likely because the generated images, despite varying in quality, mostly maintain contextual relevance to the query sentences, thereby having a similar positive impact on the inference results.

\begin{table}[h!]
\scriptsize
\centering
\begin{tabular}{@{}l|ccccc|c@{}}
\toprule
\textsc{Imagine} & $\alpha$NLI & CSQA & PIQA & SIQA & WG & Avg. \\ \midrule
Text only & 73.2 & 66.3 & 71.3 & 64.5 & 60.3 & 67.1 \\
LDM ($512\times512$) & 73.2 & 66.3 & 71.9 & 64.3 & 60.6 & 67.3 \\
DALL-E 3 ($384\times384$) & 74.5 & 66.8 & 71.9 & 64.3 & 60.6 & 67.6 \\
DALL-E 3 ($512\times512$) & \textbf{74.7} & \textbf{67.5} & \textbf{72.3} & \textbf{64.3} & \textbf{61.2} & \textbf{68.0} \\ \bottomrule
\end{tabular}%
% }
\caption{Results of using various image synthesis models for evaluation. The numbers in parentheses indicate the image resolution.}
\label{tab:img_quality}
\end{table}

\section{Impact of Adapter}
\textsc{Imagine} utilizes parallel adapters \citep{DBLP:conf/iclr/HeZMBN22-parallel_adapter} to alleviate the conflicts between the two objectives (i.e., LM, ITM) during the pre-training. In this study, we examine whether separating parameters through adapters for distinct modality objectives is truly effective. Table \ref{tab:adapter} presents the ablation results on adapters. We observe a significant decline in reasoning performance when adapters are removed. This suggests that direct training of PLMs with images adversely affects the acquisition of textual knowledge. One plausible explanation for this phenomenon is possibly related to catastrophic forgetting \citep{kirkpatrick2017overcoming}, where the model loses previously acquired knowledge (i.e., textual knowledge inherent in PLMs). This highlights the effectiveness of adapters in maintaining the model’s linguistic understanding when it learns from new modalities.

\begin{table}[t!]
\scriptsize
\centering
\begin{tabular}{@{}l|ccccc|c@{}}
\toprule
 Model & $\alpha$NLI & CSQA & PIQA & SIQA & WG & Avg. \\ \midrule
{Parallel Adapter} & \textbf{74.7} & \textbf{67.5} & \textbf{72.3} & \textbf{64.3} & \textbf{61.2} & \textbf{68.0} \\ 
{Full} & 73.0 & 65.4 & 71.1 & 61.5 & 61.2 & 66.4 \\
\bottomrule
\end{tabular}%
\caption{Evaluation results of \textsc{Imagine} with full fine-tuning (Full) and adapter tuning (Parallel Adapter). We use Synthetic VQA for pre-training.}
\label{tab:adapter}
\end{table}

\begin{table}[t!]
\scriptsize
\centering
% \resizebox{\columnwidth}{!}{%
\begin{tabular}{@{}l|ccccc|c@{}}
\toprule
RoBERTa-Large & $\alpha$NLI & CSQA & PIQA & SIQA & WG & Avg. \\ \midrule
LM & 74.3 & 65.2 & 71.9 & 62.3 & 60.5 & 66.8 \\
LM+LM & 74.3 & 66.0 & 72.1 & 64.2 & 60.4 & 67.4 \\
LM+ITM (\textsc{Imagine}) & \textbf{74.7} & \textbf{67.5} & \textbf{72.3} & \textbf{64.3} & \textbf{61.2} & \textbf{68.0} \\ \bottomrule
\end{tabular}%
% }
\caption{Results of two different ensemble methods. We use Synthetic VQA for pre-training.}
\label{tab:score ensemble}
\end{table}

To verify the effectiveness of our framework's multimodality approach, we train two unimodal models using different seeds on the Synthetic VQA dataset, utilizing only the text. We then ensemble the scores obtained from these two models. The results are presented in Table \ref{tab:score ensemble}. While ensembling scores from single modalities (LM+LM) provides performance benefits, ensembling scores from two different modalities (LM+ITM), as done in \textsc{Imagine}, proves to be the most effective. This demonstrates that the multimodality approach plays a crucial role in enhancing zero-shot reasoning performance.

\begin{figure*}[t!]
\centering
  \includegraphics[width=\linewidth]{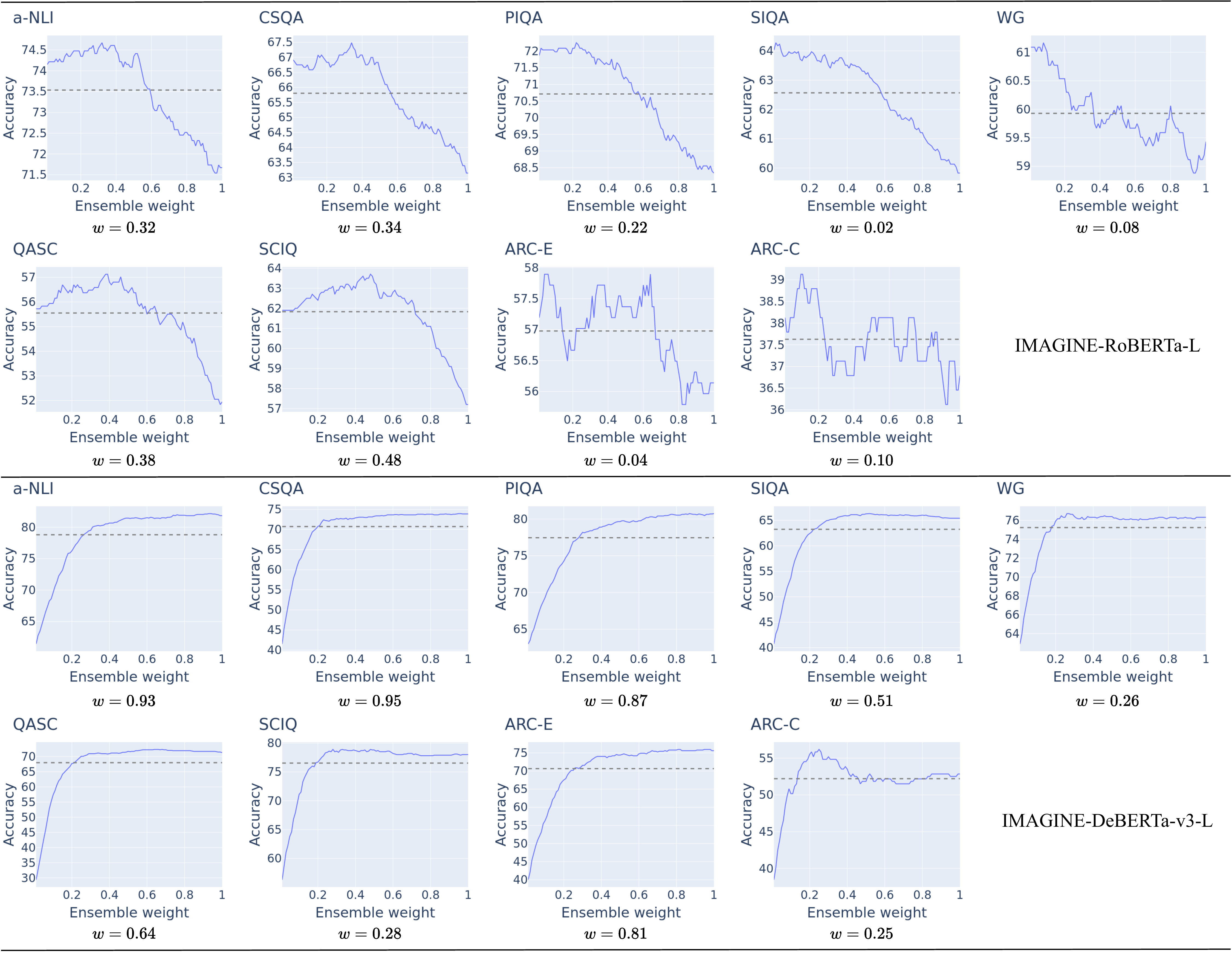}
    \caption{Model accuracy variation with different ensemble weights. The optimal $w$ for each task is shown below the plots. The line in the middle indicates the average accuracy. We use Synthetic VQA for pre-training.}
    \label{fig:ensemble_weight}
\end{figure*} 

We report the optimal ensemble weights used for our framework in Figure \ref{fig:ensemble_weight}. The larger the ensemble weight, the greater the influence of the image scores. Additionally, we draw a line indicating the average accuracy in each plot. From this, we can infer that the DeBERTa-v3-Large model utilizes image information more extensively than the RoBERTa-Large. When applying \textsc{Imagine} to DeBERTa-v3-Large, the performance improvement is greater than when using RoBERTa-Large, suggesting that visual information contributes positively to most reasoning tasks.

\section{Effect of Ensemble Inference} 
\label{app: ensemble}
\textsc{Imagine} performs reasoning by ensembling LM and ITM scores. To investigate the contributions in scores obtained from these two different modalities, we evaluate each score independently. The results are presented in Table \ref{tab:scoring}. We observe the lowest performance when evaluating only the ITM scores. However, ensembling LM scores with the ITM results in significant performance improvement across all tasks, even though the scores derived from images are much lower than those from text. This indicates that integrating machine-generated images can complement and enhance language-based reasoning abilities. 

\begin{table}[h!]
\scriptsize
\centering
% \resizebox{0.5\columnwidth}{!}{%
\begin{tabular}{@{}l|ccccc|c@{}}
\toprule
{Inference} & $\alpha$NLI & CSQA & PIQA & SIQA & WG & Avg. \\ \midrule
Ensemble & \textbf{74.7} & \textbf{67.5} & \textbf{72.3} & \textbf{64.3} & \textbf{61.2} & \textbf{68.0} \\
LM & 74.1 & 66.9 & 71.8 & 63.8 & 61.1 & 67.1 \\
ITM & 71.7 & 63.1 & 68.3 & 59.8 & 59.4 & 64.0 \\ \bottomrule
\end{tabular}%
% }
\caption{Results of the different inference strategy (LM, ITM). These strategies are evaluated on RoBERTa-L. We use Synthetic VQA for pre-training.}
\label{tab:scoring}
\end{table}

\subsection{Ablation on Training Objectives}
\label{subsec: trans}
% \paragraph{\textbf{.}}
\textsc{Imagine} employs two objectives (i.e., LM, ITM) to learn commonsense knowledge from different modalities. We perform ablations on these objectives to verify their contributions in enhancing zero-shot reasoning capabilities. Table \ref{tab:objective_ablation} shows the ablation results. Notably, omitting the LM objective leads to a significant drop in performance, underscoring the crucial role of language understanding in commonsense reasoning. Furthermore, while ITM alone does not significantly impact reasoning effectiveness, combining ITM with LM results in improved reasoning performance. These findings suggest that integrating visual information in model optimization leads to better reasoning in commonsense situations.

\begin{table}[h!]
\scriptsize
\centering
\begin{tabular}{@{}cc|ccccc|c@{}}
\toprule
LM & ITM & $\alpha$NLI & CSQA & PIQA & SIQA & WG & Avg. \\ \midrule
\checkmark & \checkmark & \textbf{74.7} & \textbf{67.5} & \textbf{72.3} & \textbf{64.3} & \textbf{61.2} & \textbf{68.0} \\
\checkmark & - & 74.3 & 65.2 & 71.9 & 62.3 & 60.5 & 66.8 \\
- & \checkmark & 71.7 & 62.0 & 68.8 & 60.0 & 59.6 & 64.4 \\
- & - & 65.6 & 45.0 & 67.6 & 47.3 & 57.5 & 56.6 \\ \bottomrule
\end{tabular}%
\caption{Ablation results on pre-training objective of \textsc{Imagine}. We use a RoBERTa-L as a backbone and Synthetic VQA for pre-training.}
\label{tab:objective_ablation}
\end{table}

\begin{figure*}[ht!]
\centering
  \includegraphics[width=\linewidth]{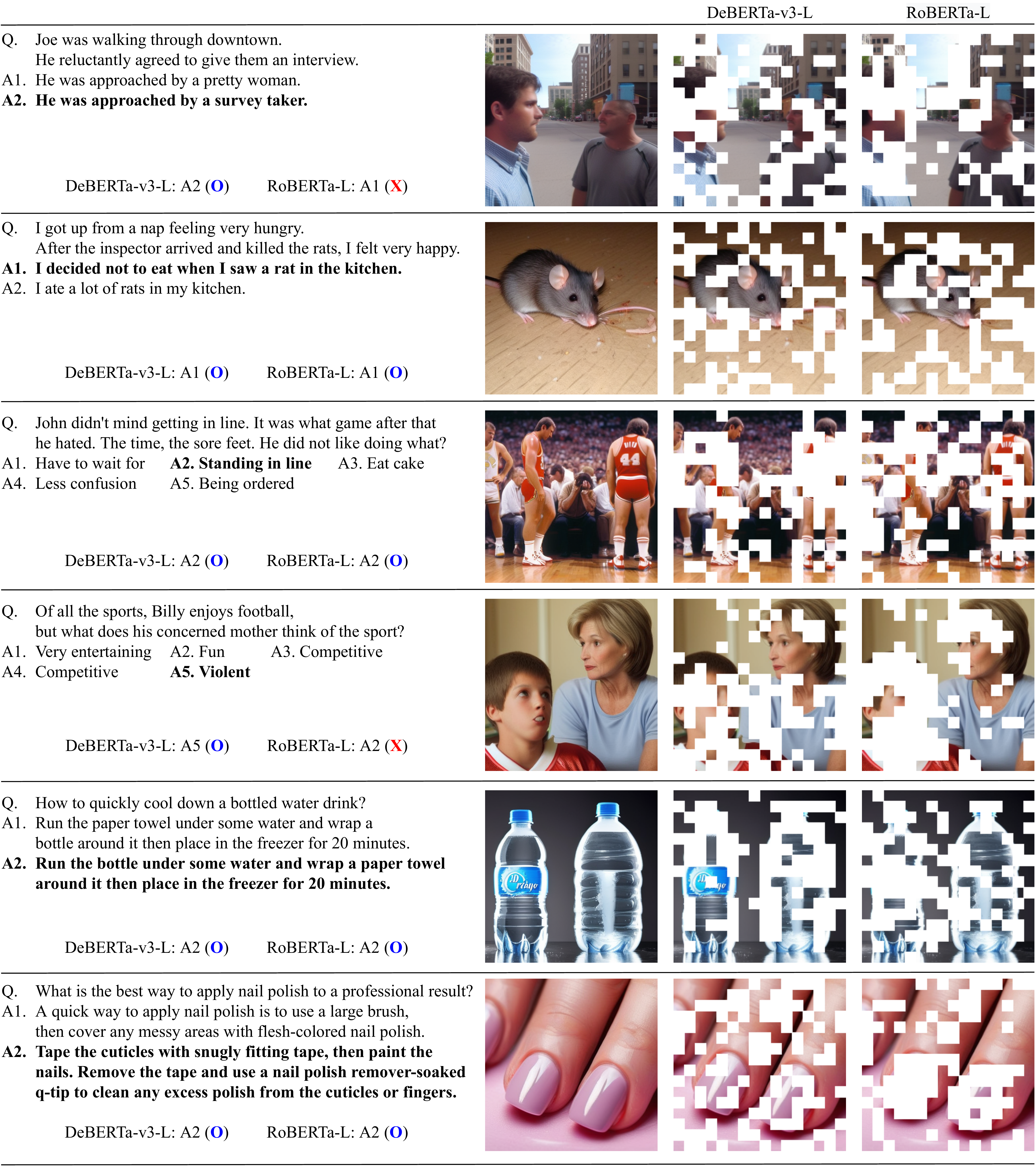}
    \caption{Randomly sampled examples from \textsc{Imagine} alongside the visualization of image attention from the Abductive NLI, CommonsenseQA, and PIQA validation sets. We use Synthetic VQA for pre-training.}
    \label{fig:ex1}
\end{figure*} 

\section{Visualization of Image Attention}
\label{app: Attention}
We aim to visualize how the model utilizes specific parts of an image. The formula to compute contextualized visual features used for computing the ITM score calculation process is similar to the attention algorithm, allowing us to derive attention scores for each image patch. Based on these scores, we erased 100 image patches with the lowest scores to understand which parts the model focused on. 
As shown in Figure \ref{fig:ex1}, each model tends to assign relatively high attention scores to objects related to the question in most cases, rather than using the image patches randomly. 
This is notable because the model can effectively capture the relationship between text and images using adapters, despite training with much less data compared to existing visual-language modeling studies \citep{DBLP:conf/icml/0008LSH23-blip2, DBLP:journals/corr/abs-2304-10592-minigpt4}. In addition, we observe that the DeBERTa-v3-Large model tends to focus more frequently on the correct parts than the RoBERTa-Large model. Figure \ref{fig:ex1} shows these cases clearly. This aligns with the result that the \textsc{Imagine} is more effective with DeBERTa-v3-Large, suggesting that a model with high generalization performance is also useful for learning new modalities.

\end{document}